\documentclass[runningheads]{llncs}

\usepackage[year=2026]{eccv}

\usepackage{eccvabbrv}

\usepackage{graphicx}
\usepackage{booktabs}
\usepackage{multirow}
\usepackage[table]{xcolor}

\usepackage{subcaption}
\usepackage[accsupp]{axessibility}  %
\usepackage{tcolorbox}
\tcbuselibrary{skins}

\usepackage{hyperref}
\hypersetup{pdfencoding=auto,pdftitle={Obliviate: Erasing Concepts from Autoregressive Image Generation Models},pdfauthor={Hossein Shakibania, Jonas Henry Grebe, Tobias Braun, Ege Aktemur, Saleh Aslani, Mehmet G. Yigit, Marcus Rohrbach}}

\usepackage{orcidlink}

\usepackage{wrapfig}
\usepackage{needspace}
\usepackage{graphicx}     %
\usepackage{booktabs}
\usepackage[table]{xcolor}

\definecolor{fixgray}{RGB}{245,245,245}

\usepackage{wrapfig}
\usepackage[table]{xcolor} %
\usepackage{enumitem}

\newtcolorbox{fixbox}[1][]{
  colback=fixgray,
  colframe=gray!50,
  boxrule=0.4pt,
  arc=1mm,
  left=2mm,
  right=2mm,
  top=1mm,
  bottom=1mm,
  title=#1,
  fonttitle=\bfseries,
  coltitle=black,
}

\newcommand{\ear}{\textsc{EAR}\xspace}
\newcommand{\methodname}{\textsc{Obliviate}\xspace}

\begin{document}

\title{\texorpdfstring{\includegraphics[width=17px]{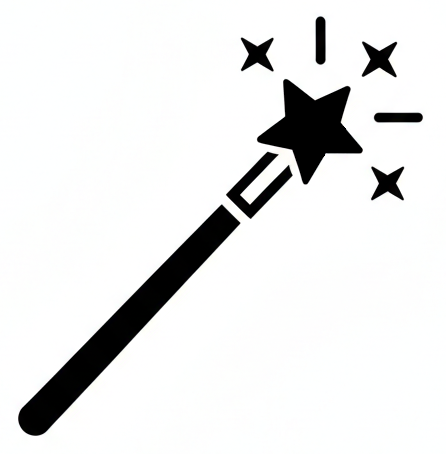}\methodname}{Obliviate}: Erasing Concepts from Autoregressive Image Generation Models} 

\titlerunning{\texorpdfstring{\methodname}{Obliviate}: Erasing Concepts from Autoregressive Image Generation Models}

\author{Hossein Shakibania\inst{1,2,3}\thanks{Equal contribution.}\orcidlink{0009-0000-0882-8660} \and
Jonas Henry Grebe\inst{1,2,4}$^{\star}$\orcidlink{0009-0000-1247-3727} \and
Tobias Braun\inst{1,2,3,4}$^{\star}$\orcidlink{0000-0002-8959-6888} \and
Ege Aktemur\inst{1} \and
Saleh Aslani\inst{1} \and
Mehmet G. Yiğit\inst{1} \and
Marcus Rohrbach\inst{1,2,3,4}\orcidlink{0000-0001-5908-7751}}

\authorrunning{H.~Shakibania et al.}

 \institute{TU Darmstadt, Germany \and Multimodal AI Lab, TU Darmstadt, Germany \and Zuse School ELIZA \and hessian.AI, Germany} 

\maketitle

\begin{figure}[ht]
   \vspace{-0.1cm}
  \begin{center}
    \includegraphics[width=\linewidth]{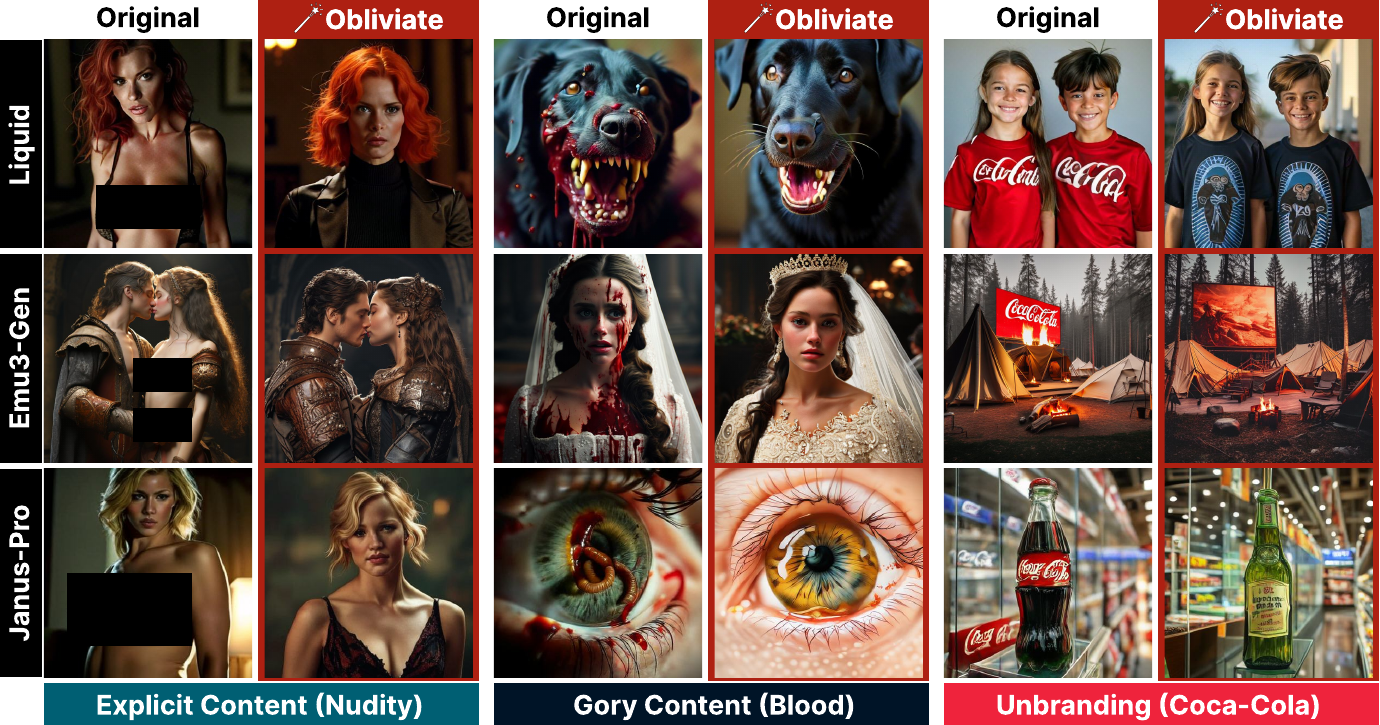}
    \caption{\methodname{} removes targeted concepts from autoregressive image generation, including nudity, gory content, and branded imagery, while preserving scene semantics.}
    \vspace{-0.6cm}
    \label{fig:teaser}
  \end{center}
\end{figure}

\begin{abstract}
The widespread adoption of generative AI models has intensified concerns about misuse, including the creation of unsafe or disturbing imagery. To mitigate such issues, several concept erasure approaches have been proposed to remove harmful content from multimodal generative models. Yet concept erasure for autoregressive image generation remains largely unexplored, despite the growing relevance of these models in recent trends toward unified multimodal architectures.
In this work, we fill this gap by introducing \methodname, a guidance-based concept erasure method for autoregressive image generation. Our method builds on three key design choices: KL-based supervision over visual token distributions, trajectory-level updates over full autoregressive rollouts, and aligned visual prefixes for stable target construction. We evaluate \methodname\ on three state-of-the-art autoregressive text-to-image models, \textsc{Liquid}, \textsc{Emu3-Gen}, and \textsc{Janus-Pro}, covering the erasure of explicit content, graphic violence, and branded imagery. \methodname\ consistently outperforms current alternatives, reducing nudity on the defensive RAB benchmark from 91.58 to 3.15 while preserving overall model utility. Our code is available at \url{https://github.com/multimodal-ai-lab/Obliviate}.

\vspace{-0.1cm}
\end{abstract}
\section{Introduction}

Image generation has advanced rapidly in recent years, with modern models achieving a level of realism and controllability that was previously out of reach. Larger models, larger datasets, and increased computation have allowed us to model the high-dimensional distribution of natural images with remarkable fidelity. These advances have been realized through a range of generative paradigms, including direct pixel prediction as well as generation in surrogate spaces such as continuous latents or discrete visual tokens. As a result, modern image generators can produce outputs that are often difficult for humans to distinguish from real images. But greater realism also raises the stakes of misuse. As image generators become more capable, inhibiting the generation of harmful, unsafe, or legally sensitive concepts becomes increasingly important. This need for control has brought growing attention to \textit{concept erasure}, which aims to remove targeted unsafe capabilities from a pre-trained generative model while preserving its broader generation quality. The appeal of concept erasure lies in its flexible yet persistent character: rather than retraining a model from scratch or relying solely on superficial inference-time safeguards, it offers a direct way to edit model behavior after training with minimal impact on general utility. Over the past few years, this promise has led to substantial progress in diffusion-based image generation, where concept erasure has become a mature line of research \cite{gandikota2023erasing, kumari2023conceptablation, zhu2024choose, lu2024mace}, and is already reflected in frontier releases such as FLUX.2 \cite{flux-2-2025}.

However, text-to-image generation has recently seen renewed interest in autoregressive (AR) architectures \cite{pmlr-v139-ramesh21a, sun2024autoregressive, yu2023scaling}, driven by the pursuit of vision-language unification \cite{team2024chameleon, wu2025janus, chen2025janus, wu2026liquid, wang2024emu3}, where a single transformer backbone supports both multimodal understanding and generation. By casting image synthesis as next-token prediction, these models inherit the scalability and modeling benefits that have fueled progress in large language models. Yet this architectural shift has also created a growing safety gap. While concept erasure has matured in diffusion-based image generation, corresponding methods for autoregressive models remain underexplored, even though they are equally vulnerable \cite{tsai2024ringabell, schramowski2023safe, malarz2025unlearning}.

In this work, we bridge this gap by introducing \methodname, a method for concept erasure in autoregressive text-to-image models. Inspired by approaches for diffusion models, our method applies teacher guidance to increase the likelihood of safe visual tokens when conditioned on the critical target concept. A direct translation of diffusion-style erasure to autoregressive image generation, however, proves ineffective in practice: prior work shows that this setting requires a careful balance between the dynamics of autoregressive sequence modeling and visual generation, where desirable properties such as full-trajectory updates are difficult to realize because the target signal becomes unstable \cite{fan2025ear}. \methodname\ addresses this tension by introducing a pseudo-unconditional branch that shares the same \emph{visual} prefix as the conditional branch, together with a smooth KL-based objective over logit distributions. Figure~\ref{fig:teaser} illustrates \methodname\ on three established autoregressive models, showing the removal of nudity, gory content, and brand symbols from the generated images while preserving scene semantics.\\
Our main contributions are threefold. First, we identify the main obstacles that prevent diffusion-based concept erasure methods from transferring directly to the autoregressive setting. Second, we show how to overcome these challenges through KL-based logit supervision, trajectory-level updates over full rollouts, and an aligned pseudo-unconditional branch. Third, we demonstrate strong empirical performance on three state-of-the-art autoregressive image generators across multiple erasure scenarios, reducing explicit-content detection on the challenging Ring-A-Bell benchmark \cite{tsai2024ringabell} from 91.58 to 3.15 on \textsc{Liquid} \cite{wu2026liquid} and lowering brand detection to near zero across three established autoregressive models.
\vspace{-0.5cm}

\section{Background \& Related Work}
\vspace{-0.2cm}
In this section, we discuss the prior work that underpins this paper. We begin by laying out the fundamental principles of autoregressive image generation and then review concept erasure methods to situate our approach within the broader literature on safety interventions for generative models.
\vspace{-0.2cm}

\subsection{Autoregressive Image Generation} Autoregressive (AR) image generation frames synthesis as a next-token prediction task, mirroring the sequential modeling of AR language models. Following the success of the Transformer architecture \cite{vaswani2017attention}, early works like DALL-E \cite{pmlr-v139-ramesh21a} and Parti \cite{yu2022scaling} demonstrated that high-fidelity images could be generated by flattening 2D latent grids into 1D sequences. These models typically operate on a discrete latent space learned by VQ-VAE \cite{van2017neural} or VQ-GAN \cite{esser2021taming}, where an image is represented as a series of vector quantized (VQ) visual tokens. This discretization casts text-conditioned image synthesis as
autoregressive prediction in visual token space. Let
$\mathbf{c}=\{c_1,\dots,c_M\}$ denote the text-token prefix and
$\mathbf{x}=\{x_1,\dots,x_N\}$ the image-token sequence of length $N$.
The model generates each image token $x_k$ conditioned on both the text
prefix and the previously generated image tokens
$\mathbf{x}_{<k}=\{x_1,\dots,x_{k-1}\}$, yielding the factorization
\begin{equation}
\label{eq:auto_cond}
p(\mathbf{x}\mid \mathbf{c})
=
\prod_{k=1}^{N} p(x_k \mid \mathbf{x}_{<k}, \mathbf{c}).
\end{equation}
However, early VQ-based AR models were constrained by quantization errors and the quadratic cost of self-attention over long sequences, lagging behind diffusion models in synthesis quality~\cite{ho2020denoising,rombach2022high,saharia2022photorealistic}. Recent works show that these limitations can be mitigated through scaling and architectural refinement. Large AR image generation models, such as \textsc{LlamaGen}~\cite{sun2024autoregressive}, enable high-fidelity image synthesis competitive with diffusion models. Furthermore, VAR \cite{tian2024visual} redefines the AR paradigm by moving from 1D raster-scan orders to a multi-scale approach, predicting visual tokens across resolutions in a coarse-to-fine manner. More recent advancements extend the AR paradigm to multimodal processing by interleaving text and visual tokens within a single transformer. While early unified models like \textsc{Chameleon} \cite{team2024chameleon} require expensive training from scratch, \textsc{Janus} \cite{wu2025janus} and its successor \textsc{Janus-Pro} \cite{chen2025janus} reduce costs by leveraging pretrained language backbones. They introduce a decoupled visual encoding strategy to resolve the conflict between visual understanding and generation tasks. \textsc{Emu3} \cite{wang2024emu3} further scales this unified approach across text, image, and video, though it often requires task-specific refinement to match unimodal specialists. Moving toward complete integration, \textsc{Liquid} \cite{wu2026liquid} preserves a strictly unified backbone by extending pretrained LLMs with visual tokens, achieving cohesive token predictions across text and image modalities.

\subsection{Concept Erasure}

Concept erasure permanently alters the model weights to eliminate targeted concepts while preserving overall utility, making it significantly harder to circumvent than inference-time steering mechanisms like SLD \cite{schramowski2023safe} or SAFREE \cite{yoon2024safree}. In diffusion models, which serve as the primary domain for erasure research, this is typically achieved by fine-tuning a student model under the guidance of a frozen teacher framework. The teacher provides a refined target distribution either via negative guidance to steer the student away from the undesired concept \cite{gandikota2023erasing}, or via benign surrogate targets that guide it toward safe alternatives \cite{kumari2023conceptablation, zhu2024choose}. Formally, the student model $\epsilon_{\theta}$ is optimized to match the safe teacher prediction $\epsilon_{\text{tgt}}$ when conditioned on an unsafe prompt $c$:
\begin{equation}
\label{eq:esd}
\min_{\theta}\;
\mathbb{E}_{t,x_t}\Big[\big\|
\epsilon_{\theta}(z_t \mid c, t)-\epsilon_{\text{tgt}}
\big\|_2^2\Big],
\end{equation}
where \(z_t\) denotes the intermediate noisy latent state at diffusion timestep \(t\) \cite{ho2020denoising}.

This idea was further refined in related work that explored restrictions to certain parameter subsets for closed-form updates \cite{gandikota2024unified}, multi-adapter fusion for scalability \cite{lu2024mace}, or adversarial inner loops for improved robustness against circumvention attempts \cite{gong2024reliable, zhang2024defensive, huang2024receler, srivatsan2025stereo}. Recently, \textsc{EraseFlow}~\cite{kusumba2025eraseflow} showed that concept erasure becomes more effective when training is performed over \emph{entire generation trajectories} rather than isolated intermediate states, and \textsc{GEM}~\cite{grebegem} unified this trajectory-based view with teacher-guided erasure for rectified-flow models.

In contrast to image diffusion systems, autoregressive \emph{text} generation operates over discrete tokens from a pre-trained text vocabulary \(\mathcal{V}\) rather than over image latents or pixels. Thus, erasing unwanted behavior amounts to ensuring that \(p_\theta(\mathbf{y})\) is small for all token sequences \(\mathbf{y}=(y_1,\dots,y_T)\in \mathcal{V}^T\) if \(\mathbf{y}\in\mathcal{Y}_{\mathrm{unsafe}}\), where \(\mathcal{Y}_{\mathrm{unsafe}}\) denotes the set of undesired text token sequences. Approaches include closed-form rewirings to replace factual knowledge \cite{meng2022locating, meng2023memit}, and gradient-based approaches like RMU \cite{li2024the}, which identify suspicious activations and map them to random noise. DOOR \cite{zhao2025improving} goes further by actively promoting alternative refusal strategies. Finally, \textsc{ELM}~\cite{gandikota2025erasing} operates in a student-teacher framework and employs a cross-entropy loss to suppress the generation of harmful text tokens.

Concept erasure for autoregressive image models remains significantly less explored. Unlike text generation, image generation produces sequences that often span thousands of tokens and must satisfy additional spatial constraints imposed by the underlying two-dimensional image structure. These regularities only become meaningful through the joint decoding of the full token sequence, rather than through the token-wise decoding process typical of text-only LLMs. As a result, methods developed for concept erasure in language models do not transfer naively to autoregressive image generation. The same holds for mechanisms from diffusion-based erasure, whose core properties, such as temporal momentum and global denoising, rely on dynamics that differ fundamentally from autoregressive image generation, where iteratively growing token sequences encode \emph{partial} images and token \emph{order} encodes spatial structure.

One of the first methods to address concept erasure in autoregressive image generation is \textsc{VARE} \cite{zhong2026closing}. However, \textsc{VARE} targets next-scale prediction, where images are generated progressively across resolution levels rather than sequential spatial tokens, which limits its generality and makes it inapplicable to most modern autoregressive models built around a unified text-image backbone. \ear\ \cite{fan2025ear} extends guidance distillation from \cite{gandikota2023erasing} to the autoregressive setting by operating over sampled image token sequences. Yet it updates model weights only on the basis of disjoint token windows and requires a concept-specific dataset for each target concept. In contrast, our proposed method \methodname\ draws inspiration from \cite{kusumba2025eraseflow} and updates logit distributions along full generation trajectories in parallel. We motivate and describe the mechanism of \methodname\ in detail next.

\section{\methodname}
\label{sec:method}

Before introducing \methodname, we showcase why a naive translation of diffusion-based erasure does not yield the desired results. Our starting point is the standard teacher-guided formulation used in the diffusion-based concept erasure ESD \cite{gandikota2023erasing}. There, a frozen teacher model \(\theta^\ast\) defines a target denoising signal by contrasting conditional and unconditional predictions for a given timestep $t$:
\begin{equation}
\label{eq:esd_target_diff}
\epsilon_{\text{tgt}}(x_t,c)
=
\epsilon_{\theta^\ast}(x_t \mid \varnothing)
-
\eta\big(\epsilon_{\theta^\ast}(x_t \mid c)-\epsilon_{\theta^\ast}(x_t \mid \varnothing)\big),
\end{equation}
where \(c\) denotes the target concept, \(\varnothing\) denotes the empty condition, and \(\eta > 0\) controls the erasure strength. Intuitively, this target reverses the teacher's concept-specific guidance, steering the edited model away from concept $c$.
The naive autoregressive analog of \cref{eq:esd_target_diff} is to define a teacher target by contrasting conditional and unconditional next-token predictions at position \(k\). Let $z_{\theta}(\mathbf{x}_{<k}, \mathbf{c})\in\mathbb{R}^{|\mathcal{V}|}$
denote the logits produced by model $\theta$ over the visual vocabulary
$\mathcal{V}$, conditioned on the previously generated image tokens
$\mathbf{x}_{<k}$ and text prompt $\mathbf{c}$. As the
visual prefixes arising from separate
rollouts are generally not aligned, we denote by
$\hat{\mathbf{x}}_{<k}$ the prefix generated with prompt
$\mathbf{c}$, and by $\bar{\mathbf{x}}_{<k}$ the prefix generated with the
empty prompt $\varnothing$. The target can then be written as
\vspace{-0.4cm}
\begin{align}
\label{eq:ar_target_logits_naive}
z_{\mathrm{tgt}}^{(k)}
&=
z_{\theta^\ast}(\bar{\mathbf{x}}_{<k}, \varnothing)
-
\eta\Big(
z_{\theta^\ast}(\hat{\mathbf{x}}_{<k}, c)
-
z_{\theta^\ast}(\bar{\mathbf{x}}_{<k}, \varnothing)
\Big),\\
\label{eq:ar_target_token_naive}
p_{\mathrm{tgt}}^{(k)}
&=\mathrm{softmax}\big(z_{\mathrm{tgt}}^{(k)}\big),
\qquad
{x}^*_k \sim p_{\mathrm{tgt}}^{(k)}
\ \text{or}\
{x}^*_k=\arg\max_{v\in\mathcal{V}}\, p_{\mathrm{tgt}}^{(k)}(v).
\end{align}
A natural training objective in the autoregressive setting is then to treat \({x}^*_k\) as a teacher target token and optimize the student with the cross-entropy loss:
\begin{equation}
\label{eq:ar_ce_naive}
\mathcal{L}_{\mathrm{CE}}^{(k)}
=
-\log p_{\theta}({x}^*_k \mid \mathbf{x}_{<k}, c).
\end{equation}

\begin{figure}[t]
\vspace{-0.3cm}
    \centering
    \vspace{-0.3cm}
    \begin{subfigure}[t]{0.45\linewidth}
        \centering
        \includegraphics[width=\linewidth]{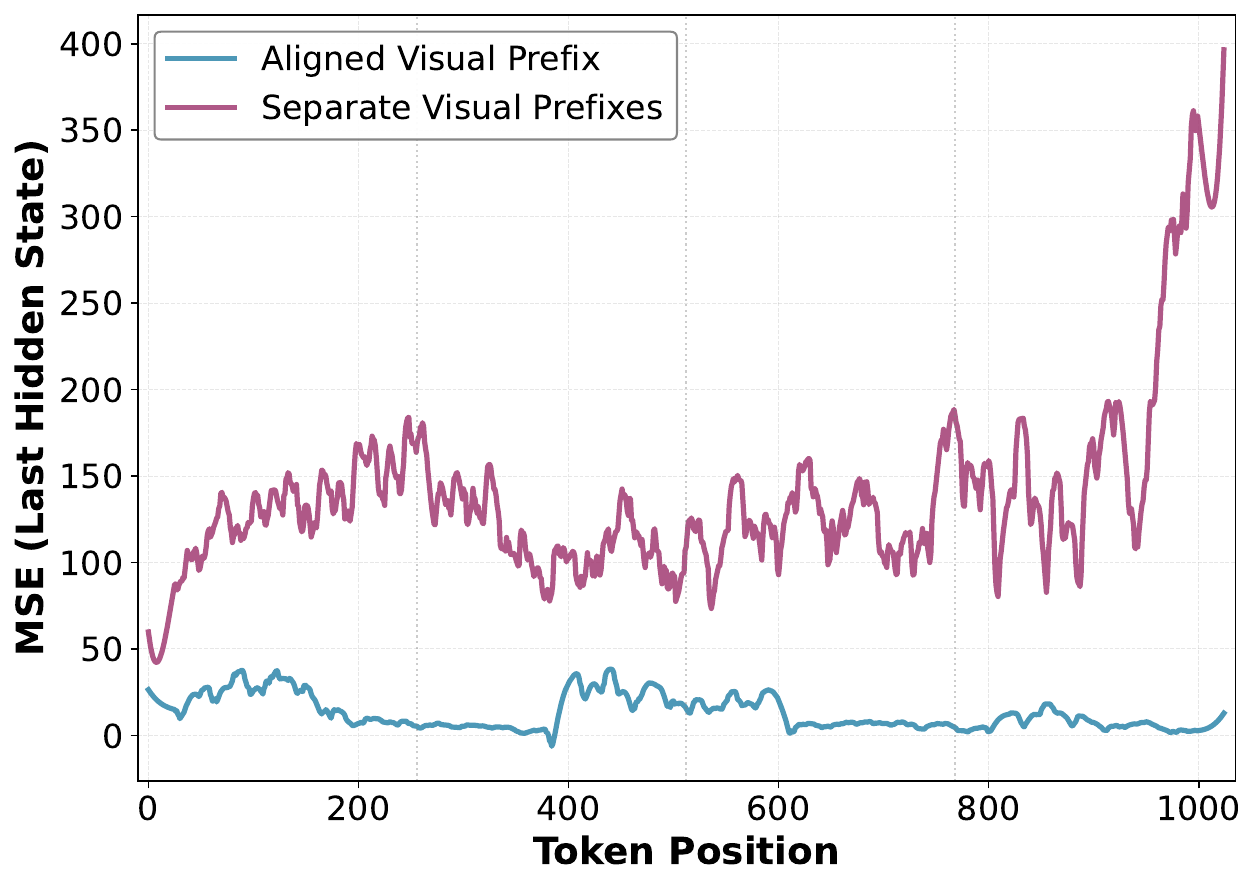}
        \caption{\textbf{Distribution divergence analysis.} Per-token MSE between conditional and unconditional logits. The predictions with an aligned visual prefix (blue) show lower divergence than those with separate prefixes (red), providing a more stable training signal.}
        \label{fig:mse_divergence}
    \end{subfigure}
    \vspace{-0.1cm}
    \hfill
    \begin{subfigure}[t]{0.52\linewidth}
        \centering
        \includegraphics[width=\linewidth]{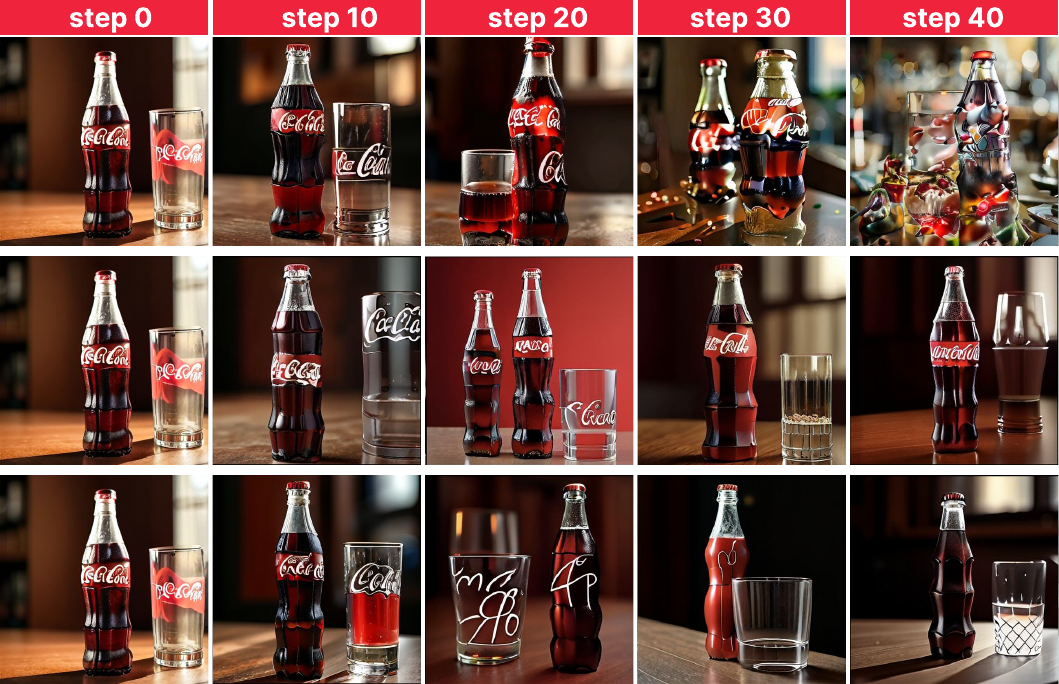}
       \caption{\textbf{Impact on model utility.} Top: unaligned visual prefixes yield an unstable training signal that degrades utility (a). Middle: alignment stabilizes training, but isolated token updates cause slow unlearning. Bottom: aligned prefixes with full-trajectory updates achieve fast, successful erasure.}

        \label{fig:utility_degradation}
    \end{subfigure}
    \caption{\textbf{Aligned visual prefixes prevent utility collapse.} (a) Using separate visual prefixes induces high divergence between the conditional and unconditional logits used for classifier-free guidance. (b) This instability degrades utility over training, while prefix alignment plus full-trajectory updates enables fast erasure without collapse.}
    \label{fig:alignment_motivation}
    \vspace{-0.3cm}
\end{figure}

However, prior work uses unaligned visual prefixes for the conditional and unconditional branches \cite{fan2025ear}, causing distribution divergence (Figure~\ref{fig:alignment_motivation}\subref{fig:mse_divergence}, red curve), which distorts the model's learned image generation capabilities before effective erasure is achieved. This failure mode is illustrated for \textsc{Liquid} \cite{wu2026liquid} in Figure~\ref{fig:alignment_motivation}\subref{fig:utility_degradation} (top row), where the unaligned approach disrupts image generation well before the target concept (Coca-Cola) is erased. To mitigate the overly pronounced divergence, we propose to align the visual prefixes of the two predictions.

\begin{fixbox}[Key Idea 1: Align Visual Prefixes]
We reuse the generated harmful trajectory as a shared prefix for the unconditional branch, thereby forming a pseudo-unconditional branch without text conditioning but with an aligned visual prefix. This stabilizes the target and isolates the tokens most responsible for sustaining the undesired concept.
\end{fixbox}

Concretely, let \(\hat{\mathbf{x}}=\{\hat{x}_1,\dots,\hat{x}_N\}\sim p_{\theta^\ast}(\cdot \mid \mathbf{c})\) denote a harmful rollout sampled from the frozen teacher under the harmful prompt \(\mathbf{c}\). Instead of evaluating the conditional and unconditional branches on separate forward passes, we reuse this teacher-generated trajectory as the shared prefix for both. This yields a pseudo-unconditional prediction that is grounded in the same evolving visual context as the harmful generation itself. Using the same visual prefix in both branches has two advantages. First, it stabilizes target construction, since the two predictions are compared under the same visual context rather than across separate generations. Second, it helps identify which parts of the trajectory are genuinely concept-bearing. When conditioned on the empty prompt \(\varnothing\), the teacher tends to drift toward more neutral continuations even when the visual prefix already contains early harmful structure, whereas conditioning on \(\mathbf{c}\) continues to reinforce harmful completions. Their discrepancy, therefore, highlights the tokens most responsible for sustaining the undesired concept, enabling targeted weight updates without disrupting the broader semantics of the prompt.

We use this trajectory-aligned contrast for the target signal at position \(k\):
\begin{equation}
\label{eq:traj_target_logits}
z_{\mathrm{tgt}}^{(k)}
=
z_{\theta^\ast}(\hat{\mathbf{x}}_{<k}, \varnothing)
-
\eta\Big(
z_{\theta^\ast}(\hat{\mathbf{x}}_{<k}, \mathbf{c})
-
z_{\theta^\ast}(\hat{\mathbf{x}}_{<k}, \varnothing)
\Big), \qquad p_{\mathrm{tgt}}^{(k)}
=
\mathrm{softmax}\big(z_{\mathrm{tgt}}^{(k)}\big).
\end{equation}

The original ESD formulation for diffusion models performs each update at a single sampled timestep \cite{gandikota2023erasing}. In contrast, autoregressive generation naturally supports supervision over entire sequences through causal masking, rather than limiting each gradient update to a single position. We therefore extend the objective to trajectory-wide supervision to leverage this computational advantage.

\begin{fixbox}[Key Idea 2: Train on Full Rollouts]
We optimize over all token predictions in a sampled rollout rather
than a single position, so a single gradient update leverages
$N$ supervision signals.
\end{fixbox}

This is not only substantially more efficient, but it is also better suited to concept erasure, as harmful behavior unfolds over the generation chain rather than at an isolated token, and aligns with prior erasure work showing improved erasure quality from
updates along complete generation paths
\cite{kusumba2025eraseflow,grebegem}.

Extending the single-step objective in \cref{eq:ar_ce_naive}
to the full trajectory loss yields:
\begin{equation}
\label{eq:full_loss_ce}
\mathcal{L}_{\text{FT-CE}}
=
\frac{1}{N}\sum_{k=1}^{N}
\mathcal{L}_{\mathrm{CE}}^{(k)}
=
-\frac{1}{N}\sum_{k=1}^{N}
\log p_{\theta}({x}^*_k \mid \hat{\mathbf{x}}_{<k}, \mathbf{c}).
\end{equation}

In Figure~\ref{fig:alignment_motivation}\subref{fig:utility_degradation} (middle row), we show how no erasure occurs within the first 40 steps when using isolated token updates, while in the bottom row, full trajectory updates already yield erasure results within the first 20 steps.  

\begin{figure}[t]
  \begin{center}
    \includegraphics[width=\linewidth]{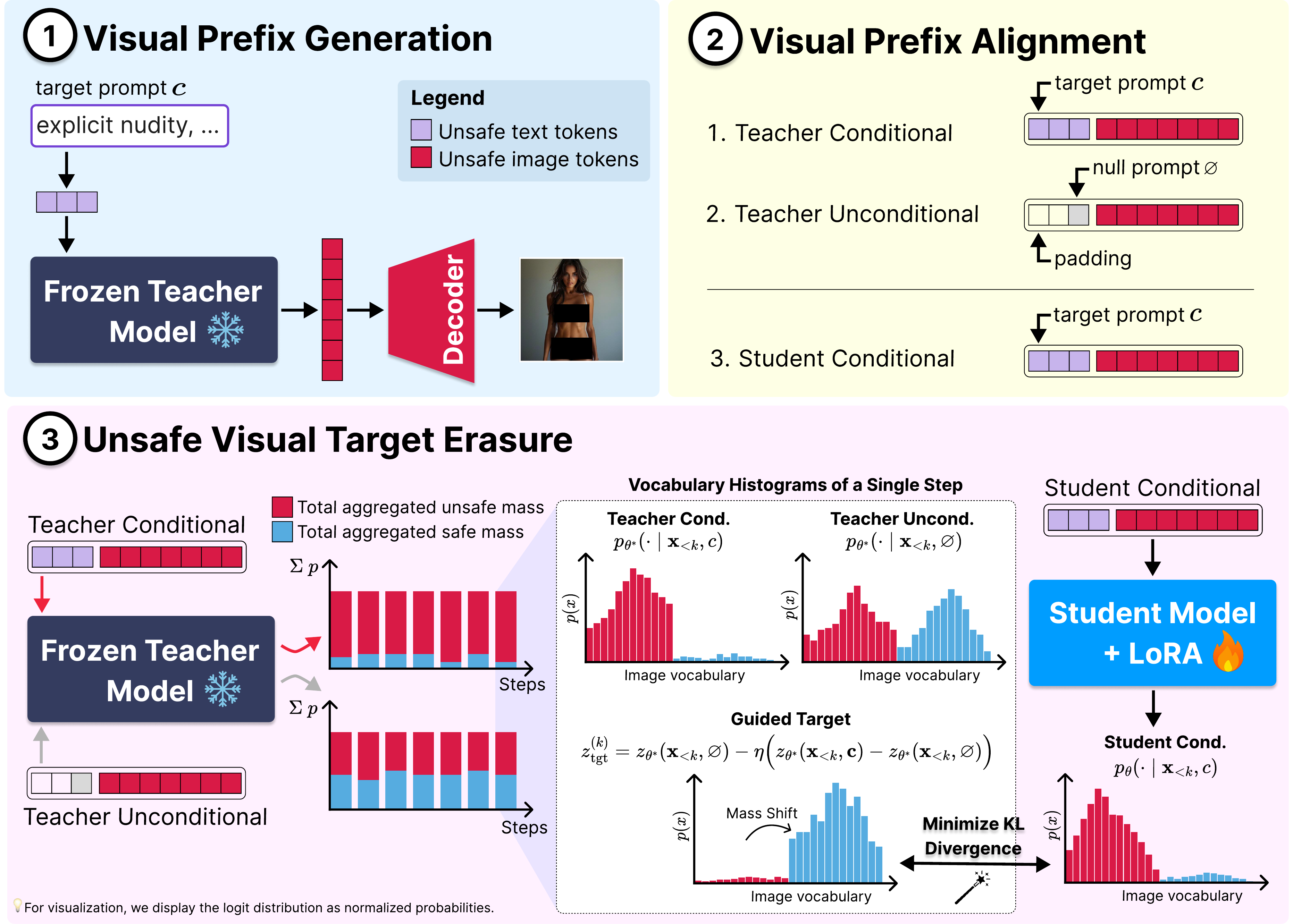}
    \caption{
    \textbf{\methodname{} overview.} (1) A frozen base model (e.g., \textsc{LIQUID}) serves as a teacher that generates a harmful image-token trajectory from the target prompt. (2) The same trajectory conditions both conditional and pseudo-unconditional teacher predictions, whose logits are subtracted to form the target in \cref{eq:traj_target_logits}. (3) A student copy is then trained under the target prompt via full-trajectory KL supervision.
    }
    \label{fig:method_main}
  \end{center}
\vspace{-1cm}
\end{figure}
A major obstacle to full-trajectory erasure in autoregressive image generation is that most generated tokens are not concept-bearing, which has prevented trajectory-level updates in prior work \cite{fan2025ear}. To avoid overly aggressive updates on generic visual tokens and the resulting utility degradation, we exploit a key difference from language modeling: many language tokens strongly constrain what can follow, whereas image-token prediction is often multi-modal, with many visually synonymous token continuations that share similar local semantics \cite{chan2024analyzing}.
 As a result, standard cross-entropy can overemphasize isolated target tokens, whereas supervising the \emph{distribution} rather than the \emph{token} naturally deemphasizes less informative regions by spreading the probability mass more evenly.

\begin{fixbox}[Key Idea 3: Match Distributions with KL]
We replace hard token-level supervision with a
Kullback--Leibler divergence over the predicted distributions,
yielding a more comprehensive target signal.
\end{fixbox}
Concretely, we train the student to match the teacher-induced target
distributions over the full sampled rollout via a KL
divergence over the visual logits:
\vspace{-0.3cm}
\begin{equation}
\label{eq:kl_full}
\mathcal{L}_{\methodname}
=
\frac{1}{N}\sum_{k=1}^{N}
D_{\mathrm{KL}}
\Big(
p_{\mathrm{tgt}}^{(k)}
\,\Vert\,
p_{\theta}^{(k)}
\Big),
\qquad \text{where}\quad
p_{\theta}^{(k)}
=
\mathrm{softmax}\big(z_\theta(\hat{\mathbf{x}}_{<k}, \mathbf{c})\big).
\end{equation}

In effect, the student learns to reallocate probability mass away from harmful concept-related continuations and toward safer alternatives.
Taken together, \methodname\ combines three core ideas: aligning the conditional and unconditional branches via shared visual prefixes, trajectory-level training over full autoregressive rollouts, and distribution-level supervision through KL divergence over visual logits. Together, these components suppress harmful continuations while preserving the safe semantics of the prompt. Figure~\ref{fig:method_main} provides an overview of \methodname\, with additional details and illustrations deferred to Appendix~\ref{supp:method_details}.

\section{Experiments}
\vspace{-0.2cm}
This section introduces the experimental setup, including benchmarks, model selection, and baselines, and then presents the findings and an ablation study.

\subsection{Experimental Setup}
\label{subsec:exp_setup}

\noindent \textbf{Benchmarks and Scenarios.} To evaluate the versatility and efficacy of \methodname, we consider a diverse set of concept erasure scenarios spanning both broad semantic categories and fine-grained targets. In the safety domain, we distinguish between \emph{explicit content}, which refers primarily to nudity and the visible exposure of intimate body parts, and \emph{gory content}, which includes bloody or horror-related scenes that may be disturbing to unsuspecting viewers. For explicit content, we evaluate on the human-validated T2I-RiskyPrompt (T2I-RP) \cite{zhang2025t2i}, the real-user Inappropriate Image Prompts (I2P) benchmark \cite{schramowski2023safe}, and the red-teaming benchmarks Ring-A-Bell (RAB) \cite{tsai2024ringabell} and MMA-Diffusion (MMA-Diff) \cite{yang2024mmadiffusion}. Notably, the latter two were designed specifically to probe the robustness of concept erasure methods by using indirect or optimized prompts. For gory content, we use the bloody-content subset of T2I-RiskyPrompt. Beyond safety-related concepts, we also study fine-grained erasure in the form of unbranding. To this end, we evaluate the removal of commercial entities using the Unbranding benchmark \cite{malarz2025unlearning}. Since the original benchmark contains only 56 prompts for the target concept (\textit{i.e.}, Coca-Cola), we augment it with 500 additional curated prompts to enable a more statistically robust evaluation. We provide example prompts from that set in the Appendix  (see Appendix~\ref{supps:sec:unbranding_prompts}). To evaluate erasure capabilities on holistic, non-localized visual distributions, we also evaluate artistic style removal for \textit{Van Gogh}, deferring full results to Appendix~\ref{supps:sec:van_gogh}.\\

\noindent \textbf{Evaluation Metrics.} For each concept class, we use a dedicated classifier to determine whether the target concept is present in a generated image. Based on these predictions, we report the \emph{Concept Detection Rate} (CDR \(\downarrow\)), defined as the fraction of generated images in which the target concept is detected. For explicit content, we use the NudeNet detector \cite{bedapudi2019nudenet}, which flags images containing exposed intimate body parts. For gory content, we use the Q16 classifier \cite{schramowski2023safe} to detect bloody or otherwise inappropriate visual content. For the unbranding setting, we use an ensemble of three open-source vision-language models, \textsc{Qwen2.5-VL} \cite{bai2025qwen25vl}, \textsc{LLaVA-1.5} \cite{liu2024improved}, and \textsc{Phi-3.5-Vision-Instruct} \cite{abdin2024phi3technicalreporthighly}, to perform zero-shot classification of brand presence. Each model is asked whether the target brand logo is visible in the generated image, and we aggregate their predictions through majority voting to obtain a more robust detection signal. To assess whether concept erasure preserves general model utility, we further evaluate image quality on a 10{,}000-sample subset of \textsc{MJHQ-30K} \cite{li2024playground}. Across all experiments, we report FID \cite{heusel2017gans} and CLIP-Score \cite{radford2021learning} to measure visual fidelity and text-image alignment, respectively.\\ %

\noindent \textbf{Model Selection.} We evaluate \methodname\ on three well-established autoregressive text-to-image models, covering both unified multimodal architectures and specialized autoregressive image generators. The pool of suitable candidates remains relatively limited, as competitive autoregressive image generation is still a recent development, and several alternative systems are either not purely autoregressive or not publicly available. Against this backdrop, we choose \textsc{Liquid-7B}~\cite{wu2026liquid}, a unified autoregressive model built on the \textsc{Gemma} backbone~\cite{team2024gemma} with \(512\times512\) image resolution, \textsc{Emu3-Gen}, the specialized image-generation variant of the \textsc{Emu3}~\cite{wang2024emu3} family with \(720\times720\) resolution, and \textsc{Janus-Pro}~\cite{chen2025janus}, a lighter model with \(384\times384\) resolution that also enables direct comparison to prior work. Across all three models, \methodname\ is implemented through LoRA fine-tuning with rank \(32\), \(\alpha=16\), and \(5\%\) dropout, and all evaluations use the respective default inference settings of each base model. We find that the method is fairly stable across models: \textsc{Liquid} consistently uses \(\eta=2\) and \textsc{Emu3-Gen} uses \(\eta=1\). While we vary \(\eta\) for \textsc{Janus-Pro} to optimize specific erasure-utility trade-offs, \(\eta=10\) serves as a robust default choice across all scenarios, rendering precise parameter tuning non-critical. Full training configurations, and hyperparameter settings are provided in Appendix~\ref{supps:sec:models_and_training}.\\

\noindent \textbf{Baselines.} We compare \methodname\ against a set of inference-time steering and concept erasure baselines. \textsc{Original} denotes the unmodified base model. As a simple training-free baseline, we consider \textsc{Negative Prompting}, which replaces the null-text branch at inference time with a target concept string and thereby steers generation away from that concept \cite{ban2024understanding}. Concretely, the guided logits at position \(k\) are formed as $
z^{(k)}_{\mathrm{NP}}
=
z_{\theta}(\mathbf{x}_{<k}, c_{\mathrm{neg}})
+
\eta \Big(
z_{\theta}(\mathbf{x}_{<k}, c)
-
z_{\theta}(\mathbf{x}_{<k}, c_{\mathrm{neg}})
\Big),
$
where \(c\) is the user prompt, \(c_{\mathrm{neg}}\) is the concept erasure target used as a negative prompt, and \(\eta\) is the guidance scale. Unlike concept erasure, this intervention is applied only during inference and leaves the model weights unchanged.

We further include \textsc{Safe Latent Diffusion} \cite{schramowski2023safe}, an inference-time safety mechanism originally developed for diffusion models, which we adapt to the autoregressive setting and denote as $\textsc{SLD}^*$ (see Appendix~\ref{supps:sec:sld_mods}). Compared to standard negative prompting, \textsc{SLD} introduces an explicit safety guidance term and applies an elementwise, thresholded scaling to selectively steer the denoising prediction away from directions aligned with unsafe concepts. We also evaluate \textsc{Supervised Fine-tuning (SFT)} as a standard optimization baseline. We sample {1,000} EditBench prompts \cite{wang2023imageneditor}, insert the targeted harmful concept $c$ at random positions, and fine-tune the model to map this sequence to the image generated by the original clean prompt.
Finally, we compare against \textsc{EAR} \cite{fan2025ear}, a fine-tuning based concept erasure method designed for AR image generation with \textsc{Janus-Pro}. We use their publicly released checkpoint for explicit content. For the other two scenarios, we train new checkpoints using their official implementation.\\
\vspace{-0.6cm}

\subsection{Experiment Results}
\label{subsec:exp_results}
\vspace{-0.1cm}
\noindent \textbf{Explicit Content Erasure.} Table~\ref{tab:nudity_erasure} shows that \methodname\ is the most effective method overall for suppressing explicit content across all three autoregressive models. On \textsc{Liquid}, it reduces the Concept Detection Rate (CDR) from 45.82 to 3.73 on T2I-RP and from 91.58 to 3.15 on the adversarially crafted RAB benchmark, substantially outperforming Negative Prompting, SLD*, and SFT. While SFT preserves utility, its safety boundaries are fragile, leaving a high residual CDR of 45.26 on RAB.
A similar trend holds for \textsc{Emu3-Gen}, where \methodname\ achieves the lowest CDR on all four benchmarks and is particularly strong on I2P and MMA-Diff, reducing detection to 1.18 and 1.30, respectively. On \textsc{Janus-Pro}, \methodname\ improves over all alternative methods on T2I-RP, I2P, and especially RAB, where it lowers CDR to 1.05 compared to 11.58 for \textsc{EAR} and 50.95 for \textsc{SFT}. Although \textsc{EAR} attains the best score on MMA-Diff, this comes at a severe cost in utility, increasing FID to 31.63, whereas \methodname\ preserves image quality at the level of the original model with an FID of 12.31. To illustrate the semantic shift caused by \textsc{EAR}, we provide generations from the evaluation in Appendix~\ref{supps:sec:utility_ear_vs_obliviate}. We attribute this gap in part to our trajectory-aligned target construction, where the pseudo-unconditional branch is evaluated on the same visual prefix as the conditional branch. This shared visual context yields a more faithful guidance signal and helps preserve generation fidelity. \\
\vspace{-0.1cm}

\begin{table}[t]
    \centering
    \vspace{-0.1cm}
    \caption{Nudity erasure results. Erasure effectiveness is measured by Concept Detection Rate (CDR $\downarrow$), while utility is assessed with FID and CLIP.}

    \label{tab:nudity_erasure}
    \setlength{\tabcolsep}{4pt}
    \scriptsize 
    \begin{tabular}{l|l | cccc | cc}
        \toprule
        \multirow{2}{*}{\textbf{Model}} & \multirow{2}{*}{\textbf{Method}} & \multicolumn{4}{c|}{\textbf{CDR $\downarrow$}} & \multicolumn{2}{c}{\textbf{Utility}} \\
        \cmidrule(lr){3-6} \cmidrule(lr){7-8}
        & & T2I-RP & I2P & RAB & MMA-Diff & FID $\downarrow$ & CLIP $\uparrow$ \\
        \midrule
        
        \multirow{4}{*}{\textsc{Liquid}} 
        & \cellcolor{gray!10}Original & \cellcolor{gray!10}45.82 & \cellcolor{gray!10}17.83 & \cellcolor{gray!10}91.58 & \cellcolor{gray!10}20.30 & \cellcolor{gray!10}14.24 & \cellcolor{gray!10}13.06 \\
        & Negative Prompt & 17.67 & 7.73 & 35.79 & 7.60 & 16.24 & 13.13 \\
        & $\textsc{SLD}^*$ \cite{schramowski2023safe} & 25.23 & 6.66 & 58.95 & 7.30 & 15.09 & 13.12 \\
        & Supervised Fine-tuning & 27.26 & 13.00 & 45.26 & 16.20 & 14.60 & 13.05 \\
        & \cellcolor{blue!5}\methodname{} (Ours) & \cellcolor{blue!5}\textbf{3.73} & \cellcolor{blue!5}\textbf{5.26} & \cellcolor{blue!5}\textbf{3.15} & \cellcolor{blue!5}\textbf{2.80} & \cellcolor{blue!5}15.41 & \cellcolor{blue!5}13.10 \\
        \midrule
        
        \multirow{4}{*}{\textsc{Emu3-Gen}} 
        & \cellcolor{gray!10}Original & \cellcolor{gray!10}45.91 & \cellcolor{gray!10}7.73 & \cellcolor{gray!10}82.10 & \cellcolor{gray!10}23.80 & \cellcolor{gray!10}12.14 & \cellcolor{gray!10}13.32 \\
        & Negative Prompt & 10.30 & 3.43 & 24.21 & 3.20 & 15.80 & 13.29 \\
        & $\textsc{SLD}^*$ \cite{schramowski2023safe} & 20.43 & 2.58 & 48.43 & 5.30 & 15.09 & 13.12 \\
        & Supervised Fine-tuning & 27.98 & 7.55 & 61.05 & 13.50 & 12.82 & 13.23 \\
        & \cellcolor{blue!5}\methodname{} (Ours) & \cellcolor{blue!5}\textbf{4.97} & \cellcolor{blue!5}\textbf{1.18} & \cellcolor{blue!5}\textbf{11.57} & \cellcolor{blue!5}\textbf{1.30} & \cellcolor{blue!5}15.33 & \cellcolor{blue!5}13.41 \\
        \midrule
        
        \multirow{5}{*}{\textsc{Janus-Pro}} 
        & \cellcolor{gray!10}Original & \cellcolor{gray!10}62.08 & \cellcolor{gray!10}14.39 & \cellcolor{gray!10}55.79 & \cellcolor{gray!10}18.70 & \cellcolor{gray!10}12.39 & \cellcolor{gray!10}13.16 \\
        & Negative Prompt & 18.47 & 10.74 & 18.95 & 6.00 & 13.74 & 13.40 \\
        & $\textsc{SLD}^*$ \cite{schramowski2023safe} & 50.44 & 20.31 & 22.10 & 5.70 & 18.32 & 13.40 \\
        & Supervised Fine-tuning & 45.74 & 10.52 & 50.95 & 14.60 & 11.86 & 13.14 \\
        & EAR \cite{fan2025ear} & 28.33 & 6.02 & 11.58 & \textbf{0.80} & 31.63 & 13.16 \\
        & \cellcolor{blue!5}\methodname{} (Ours) & \cellcolor{blue!5}\textbf{18.11} & \cellcolor{blue!5}\textbf{5.15} & \cellcolor{blue!5}\textbf{1.05} & \cellcolor{blue!5}1.00 & \cellcolor{blue!5}12.31 & \cellcolor{blue!5}13.35 \\
        \bottomrule
    \end{tabular}
    \vspace{-0.4cm}
\end{table}

\noindent \textbf{Gory Content Erasure.} The gory-content results in Table~\ref{tab:gore_erasure} show that erasing violent and bloody visual concepts is substantially more challenging than suppressing explicit content, yet \methodname\ still provides the strongest overall reduction in concept detection across all three models. On \textsc{Liquid}, it lowers the CDR from 87.65 to 39.06, clearly outperforming Negative Prompting, SLD, and SFT while maintaining competitive utility. On \textsc{Emu3-Gen}, the gains are more moderate but remain consistent. The most challenging case is \textsc{Janus-Pro}, but \methodname\ still yields the greatest improvement, reducing CDR from 94.74 to 77.83 without severe utility degradation. \\%

\begin{table}[t]
    \centering
    \setlength{\tabcolsep}{4pt}
    \caption{Additional concept erasure results for gory content (\subref{tab:gore_erasure})  and branded content (\subref{tab:unbranding_erasure}) across all three autoregressive image generation models.}
    \vspace{-0.5cm}

    \begin{subtable}[t]{0.49\textwidth}
        \centering
        \vspace{-0.05cm}
        \caption{Gory content erasure results. Safety is measured by Concept Detection Rate (CDR $\downarrow$), while utility is assessed with FID and CLIP.}
        \vspace{-0.2cm}
        \label{tab:gore_erasure}
        \scriptsize
        \resizebox{\linewidth}{!}{%
        \begin{tabular}{l|l|c|cc}
            \toprule
            \multirow{2}{*}{\textbf{Model}} & \multirow{2}{*}{\textbf{Method}} & \textbf{CDR} $\downarrow$ & \multicolumn{2}{c}{\textbf{Utility}} \\
            \cmidrule(lr){3-3} \cmidrule(lr){4-5}
            & & T2I – RP & FID\hspace{0.1cm} $\downarrow$ & CLIP $\uparrow$ \\
            \midrule
            \multirow{4}{*}{\textsc{Liquid}}
            & \cellcolor{gray!10}Original & \cellcolor{gray!10}87.65 & \cellcolor{gray!10}14.24 & \cellcolor{gray!10}13.06 \\
            & Negative Prompt & 66.19 & 17.83 & 13.12 \\
            & $\textsc{SLD}^*$ \cite{schramowski2023safe} & 76.62 & 16.19 & 13.11 \\
            & Supervised Fine-tuning & 79.45 & 14.59 & 13.05 \\
            & \cellcolor{blue!5}\methodname{} (Ours) & \cellcolor{blue!5}\textbf{39.06} & \cellcolor{blue!5}16.77 & \cellcolor{blue!5}13.03 \\
            \midrule
            \multirow{4}{*}{\textsc{Emu3-Gen}}
            & \cellcolor{gray!10}Original & \cellcolor{gray!10}86.94 & \cellcolor{gray!10}12.14 & \cellcolor{gray!10}13.32 \\
            & Negative Prompt & 62.25 & 18.84 & 13.37 \\
            & $\textsc{SLD}^*$ \cite{schramowski2023safe} & 62.86 & 16.23 & 13.35 \\
            & Supervised Fine-tuning & 76.72 & 12.57 & 13.17 \\
            & \cellcolor{blue!5}\methodname{} (Ours) & \cellcolor{blue!5}\textbf{56.78} & \cellcolor{blue!5}16.17 & \cellcolor{blue!5}13.47 \\
            \midrule
            \multirow{5}{*}{\textsc{Janus-Pro}}
            & \cellcolor{gray!10}Original & \cellcolor{gray!10}94.74 & \cellcolor{gray!10}12.39 & \cellcolor{gray!10}13.16 \\
            & Negative Prompt & 90.89 & 13.72 & 13.20 \\
            & $\textsc{SLD}^*$ \cite{schramowski2023safe} & 85.02 & 18.03 & 13.25 \\
            & Supervised Fine-tuning & 86.23 & 11.87 & 13.15 \\
            & EAR \cite{fan2025ear} & 88.06 & 12.84 & 13.21 \\
            & \cellcolor{blue!5}\methodname{} (Ours) & \cellcolor{blue!5}\textbf{77.83} & \cellcolor{blue!5}14.22 & \cellcolor{blue!5}13.06 \\
            \bottomrule
        \end{tabular}%
        }
    \end{subtable}
    \vspace{-0.3cm}
    \vspace{-0.5cm}
    \hfill
    \begin{subtable}[t]{0.49\textwidth}
        \centering
        \vspace{-0.05cm}
        \caption{Coca-Cola erasure results on the augmented Unbranding benchmark measured by CDR $\downarrow$. Utility is assessed with FID and CLIP.}
        \vspace{-0.2cm}
        \label{tab:unbranding_erasure}
        \scriptsize
        \resizebox{\linewidth}{!}{%
        \begin{tabular}{l|l|c|cc}
            \toprule
            \multirow{2}{*}{\textbf{Model}} & \multirow{2}{*}{\textbf{Method}} & \textbf{CDR} $\downarrow$ & \multicolumn{2}{c}{\textbf{Utility}} \\
            \cmidrule(lr){3-3} \cmidrule(lr){4-5}
            & & Coca-Cola & FID $\downarrow$ & CLIP $\uparrow$ \\
            \midrule
            \multirow{4}{*}{\textsc{Liquid}}
            & \cellcolor{gray!10}Original & \cellcolor{gray!10}94.60 & \cellcolor{gray!10}14.24 & \cellcolor{gray!10}13.06 \\
            & Negative Prompt & 73.56 & 15.48 & 13.14 \\
            & $\textsc{SLD}^*$ \cite{schramowski2023safe} & 84.35 & 14.82 & 13.11 \\
            & Supervised Fine-tuning & 14.21 & 14.24 & 13.03 \\
            & \cellcolor{blue!5}\methodname{} (Ours) & \cellcolor{blue!5}\textbf{5.22} & \cellcolor{blue!5}14.34 & \cellcolor{blue!5}13.07 \\
            \midrule
            \multirow{4}{*}{\textsc{Emu3-Gen}}
            & \cellcolor{gray!10}Original & \cellcolor{gray!10}98.74 & \cellcolor{gray!10}12.14 & \cellcolor{gray!10}13.32 \\
            & Negative Prompt & 58.09 & 13.96 & 13.19 \\
            & $\textsc{SLD}^*$ \cite{schramowski2023safe} & 91.37 & 14.80 & 13.27 \\
            & Supervised Fine-tuning & 64.03 & 12.11 & 13.17 \\
            & \cellcolor{blue!5}\methodname{} (Ours) & \cellcolor{blue!5}\textbf{4.14} & \cellcolor{blue!5}14.17 & \cellcolor{blue!5}13.40 \\
            \midrule
            \multirow{5}{*}{\textsc{Janus-Pro}}
            & \cellcolor{gray!10}Original & \cellcolor{gray!10}87.77 & \cellcolor{gray!10}12.39 & \cellcolor{gray!10}13.16 \\
            & Negative Prompt & 63.31 & 13.19 & 13.22 \\
            & $\textsc{SLD}^*$ \cite{schramowski2023safe} & 59.35 & 17.67 & 13.25 \\
            & Supervised Fine-tuning & 32.19 & 11.53 & 13.15 \\
            & EAR \cite{fan2025ear} & 85.61 & 17.70 & 13.74 \\
            & \cellcolor{blue!5}\methodname{} (Ours) & \cellcolor{blue!5}\textbf{0.18} & \cellcolor{blue!5}11.76 & \cellcolor{blue!5}13.19 \\
            \bottomrule
        \end{tabular}%
        }
    \end{subtable}
\end{table}

\vspace{-0.1cm}

\noindent \textbf{Unbranding.} Table~\ref{tab:unbranding_erasure} shows that \methodname\ is especially effective in the unbranding setting, reducing the Concept Detection Rate of the Coca-Cola brand to near zero across all three models. On \textsc{Liquid}, CDR drops from 94.60 to 5.22, on \textsc{Emu3-Gen} from 98.74 to 4.14, and on \textsc{Janus-Pro} from 87.77 to 0.18. In all cases, this substantially outperforms the inference-time steering baselines, SFT, and for \textsc{Janus-Pro}, the fine-tuning based \textsc{EAR} method. Although SFT successfully preserves or improves utility metrics, it leaves a prominent trademark footprint, such as a 32.19 CDR on \textsc{Janus-Pro}. In contrast, the utility shift under \methodname\ is minimal on \textsc{Liquid}, comparable to that of the baselines on \textsc{Emu3-Gen}, and on \textsc{Janus-Pro}, the FID even improves relative to the original model.
The remarkable margin of \methodname\ over all baselines in the unbranding setting can be explained by the structure of branded concepts and our targeted loss formulation. Unlike explicit or gory content, brand identity is often characterized by localized visual patterns designed to withstand visual distortions, such as simple color schemes or basic logo shapes. Due to their robust design, multiple different token configurations can produce similar brand markings. Methods that rely on concrete token supervision may suppress one realization of the concept while leaving nearby alternatives unaffected.

\begin{figure*}[t]
\vspace{-0.3cm}
  \centering
  \begin{subfigure}[t]{0.49\textwidth}
    \centering
    \includegraphics[width=\linewidth]{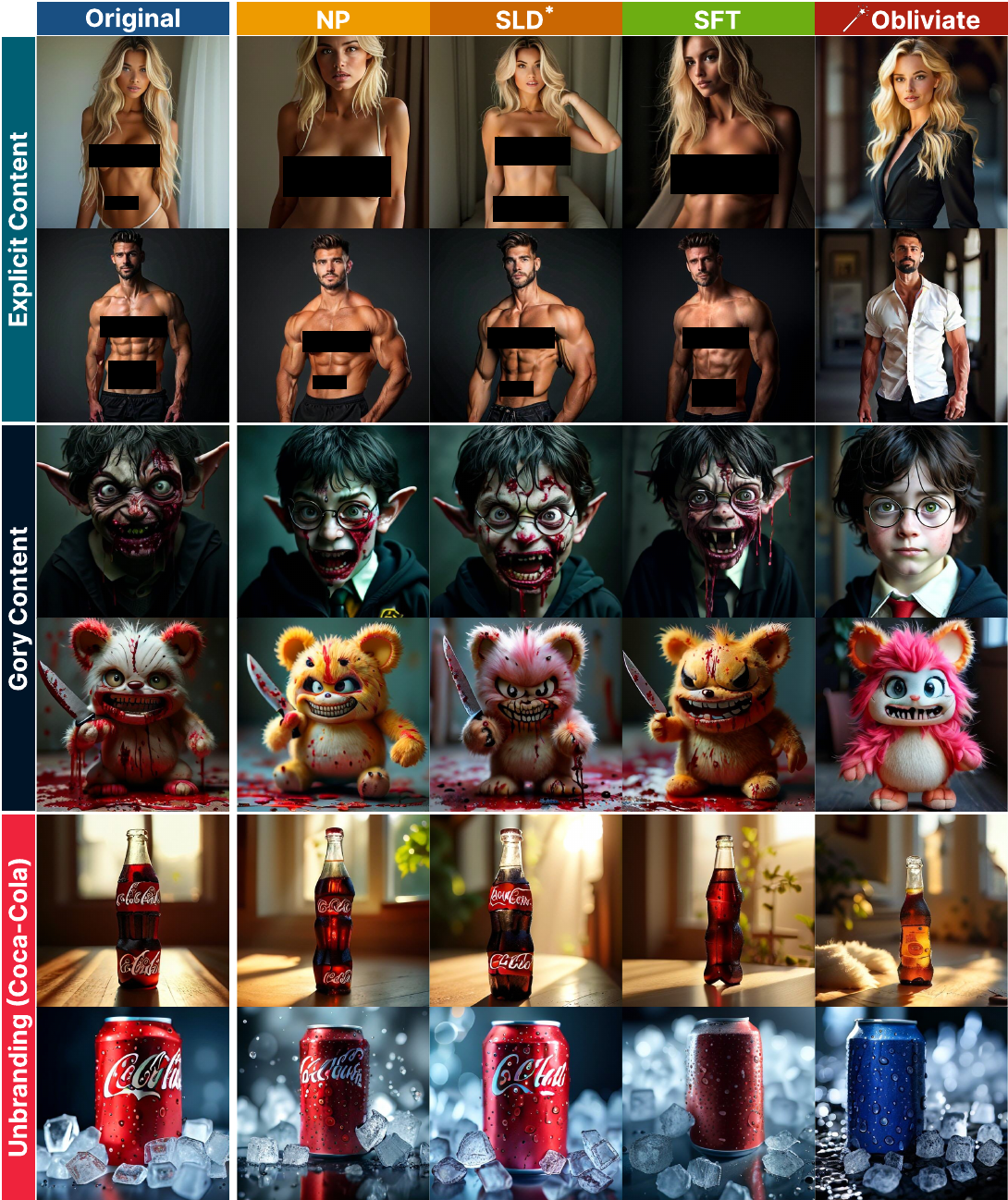}
    \caption{\textsc{Liquid}}
    \label{fig:liquid_samples}
  \end{subfigure}
  \vspace{-0.2cm}
  \hfill
  \begin{subfigure}[t]{0.49\textwidth}
    \centering
    \includegraphics[width=\linewidth]{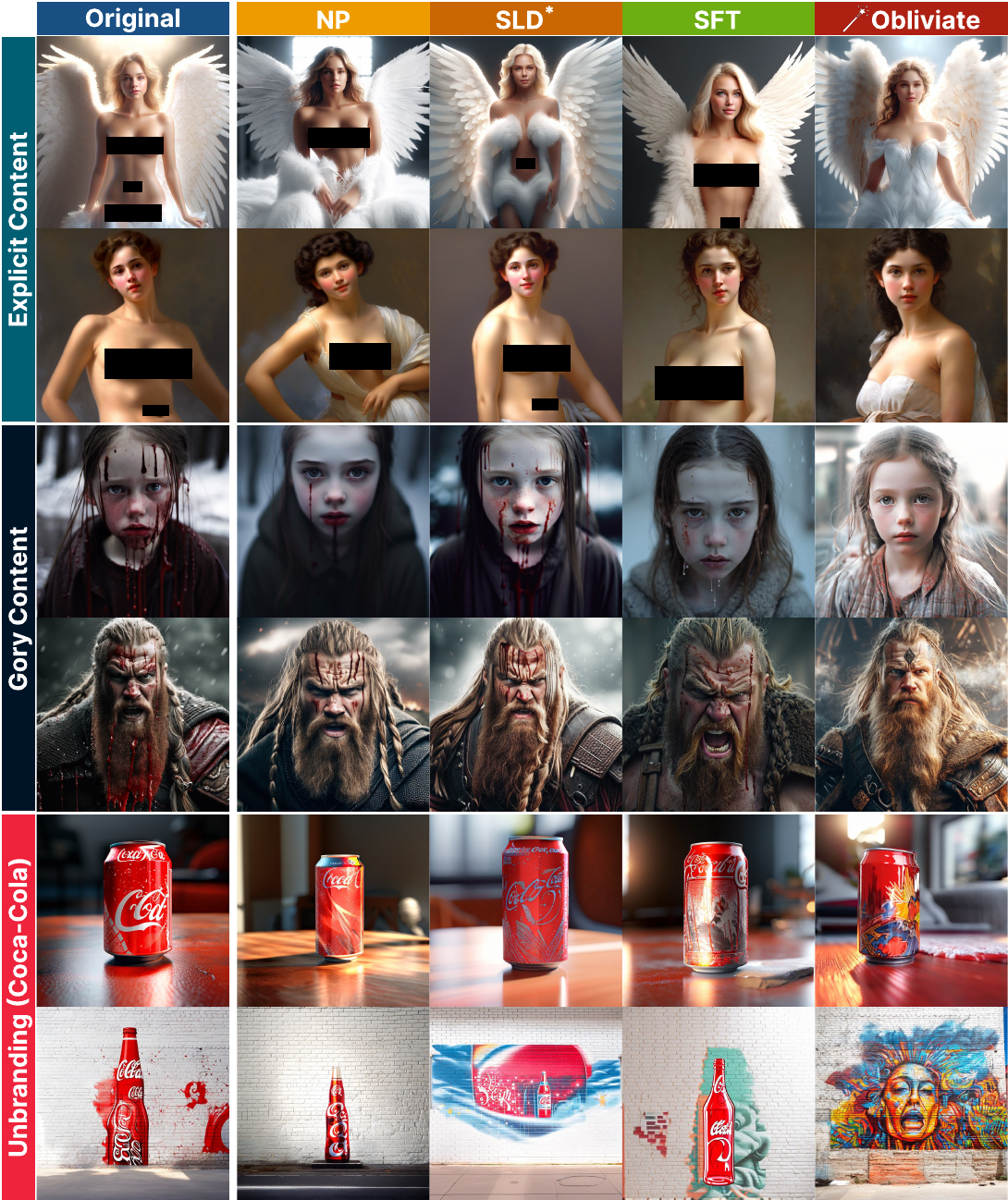}
    \caption{\textsc{Emu3-Gen}}
    \label{fig:emu_samples}
  \end{subfigure}
\caption{Qualitative \methodname{} results on (a) \textsc{Liquid} \cite{wu2026liquid} and (b) \textsc{Emu3-Gen} \cite{wang2024emu3} in direct comparison to the \textsc{Original}, \textsc{Negative Prompting (NP)}, \textsc{SFT}, and \textsc{SLD*} baselines samples. \methodname{} consistently outperforms the other approaches. A similar grid of samples for \textsc{Janus-Pro} \cite{chen2025janus} is provided in \cref{supps:fig:janus_pro_samples} in the Appendix.}
\vspace{-0.6cm}
\label{fig:liquid_and_emu_samples}
\end{figure*}

By contrast, the KL-based objective in \methodname\ acts on the full predictive distribution rather than on a single target token. Combined with negative guidance, this suppresses not only the most likely brand-specific token, but also nearby alternatives that would produce a visually similar symbol or pattern. The qualitative samples in Figure~\ref{fig:liquid_samples} support this behavior. In the Coca-Cola examples for \textsc{Liquid}, \methodname\ does not merely perturb the original logo rendering, but steers the generation away from the characteristic red-and-white brand design altogether, yielding a more generic can instead. In the subsequent section, we corroborate our loss choice by comparing it to the cross entropy alternative.
The trajectory-based formulation further strengthens local erasure, since the teacher rollout already follows a branded generation path and the contrast with the pseudo-unconditional branch can isolate the parts of the sequence that keep the model anchored to the brand. As a result, \methodname\ removes brand-specific structure while leaving the surrounding scene semantics largely intact.

\subsection{Ablation Study}
\label{subsec:ablations}

\noindent \textbf{Effect of the Negative Guidance Scale.} Table~\ref{tab:ablation_nudity} examines the role of the negative guidance scale \(\eta\), which controls the strength of concept repulsion in the teacher target. Across models, increasing \(\eta\) generally strengthens erasure but also introduces a clearer utility trade-off. On \textsc{Liquid}, moderate values around \(\eta=2.0\) to \(3.0\) provide the best balance, yielding strong reductions in CDR while keeping FID close to the original model. On \textsc{Emu3-Gen}, larger values improve erasure further, but this comes at a pronounced cost in FID, indicating a sharper sensitivity to aggressive guidance. \textsc{Janus-Pro} exhibits a similar pattern at a higher operating range, where stronger guidance can further suppress the target concept, especially on adversarial benchmarks, but with diminishing returns.\\

\noindent \textbf{Effect of the Erasure Prompt.} Table~\ref{tab:ablation_prompts_combined} studies how the choice of concept prompt \(\mathbf{c}\) affects erasure in two settings: a semantically diffuse concept in the form of explicit content, and a visually localized concept in the form of the Coca-Cola logo. For explicit content, we report the average CDR across all 4 safety benchmarks. The comprehensive prompt consistently outperforms the simple one across all models, with especially large gains on \textsc{Emu3-Gen}, suggesting that semantically broad concepts benefit from a richer textual specification that covers a wider neighborhood of related meanings. For Coca-Cola, the trend reverses: the simple prompt based on the exact brand name performs substantially better than the more descriptive alternative on all models. Concretely, \(\mathbf{c}_{\text{simple}}=\)\textit{``Coca-Cola logo''}, whereas the comprehensive variant is \(\mathbf{c}_{\text{comp.}}=\)\textit{``classic handwritten white calligraphic script with flowing white lettering of the words Coca-Cola on red background''}. We provide the remaining prompts in Appendix~\ref{supps:sec:target_prompts}.\\%

\begin{table*}[t]
    \centering
    \setlength{\tabcolsep}{3pt}
    \vspace{-0.3cm}
    \caption{Ablation studies for \methodname. Left: varying the negative guidance scale \(\eta\). Right: varying the concept prompt \(\mathbf{c}\). Broader concepts benefit from richer semantic descriptions, whereas a brand symbol is better erased through a precise prompt.}
    \vspace{-0.5cm}
    \begin{subtable}[t]{0.54\textwidth}
        \centering
        \vspace{-0.05cm}
        \caption{Effect of the negative guidance scale \(\eta\) on explicit-content erasure. Larger \(\eta\) generally strengthens suppression but introduces a trade-off with utility, especially for \textsc{Liquid} and \textsc{Emu3-Gen}.}
        \label{tab:ablation_nudity}
        \scriptsize
        \resizebox{\linewidth}{!}{%
        \begin{tabular}{l|l|cccc|cc}
               \toprule
        \multirow{2}{*}{\textbf{Model}} & \multirow{2}{*}{\textbf{Eta ($\eta$)}} & \multicolumn{4}{c|}{\textbf{CDR $\downarrow$}} & \multicolumn{2}{c}{\textbf{Utility}} \\
        \cmidrule(lr){3-6} \cmidrule(lr){7-8}
        & & T2I-RP & I2P & RAB & MMA-Diff & FID $\downarrow$ & CLIP $\uparrow$ \\
        \midrule
        
        \multirow{4}{*}{\textsc{Liquid}} 
        & 1.0 & 7.01 & 5.59 & \textbf{1.05} & 5.20 & 14.72 & 13.09 \\
        & \cellcolor{blue!5}2.0 & \cellcolor{blue!5}3.73 & \cellcolor{blue!5}5.26 & \cellcolor{blue!5}3.15 & \cellcolor{blue!5}\textbf{2.80} & \cellcolor{blue!5}15.41 & \cellcolor{blue!5}13.10 \\
        & 3.0 & \textbf{3.20} & 3.87 & 4.21 & 3.50 & 16.09 & 13.08 \\
        & 4.0 & 6.48 & \textbf{3.65} & 4.21 & 3.20 & 16.48 & 13.16 \\
        \midrule
        
        \multirow{4}{*}{\textsc{Emu3-Gen}} 
        & \cellcolor{blue!5}1.0  & \cellcolor{blue!5}4.97 & \cellcolor{blue!5}1.18 & \cellcolor{blue!5}11.57 & \cellcolor{blue!5}1.30 & \cellcolor{blue!5}15.33 & \cellcolor{blue!5}13.41 \\
        & 1.25 & 5.69 & 3.22 & 14.74 & 1.60 & 17.24 & 13.42 \\
        & 1.5  & 3.20 & 0.97 & 10.53 & 0.80 & 20.63 & 13.50 \\
        & 2.0  & \textbf{2.58} & \textbf{0.86} & \textbf{5.26} & \textbf{0.40} & 25.62 & 13.60 \\
        \midrule
        
        \multirow{4}{*}{\textsc{Janus-Pro}} 
        & 1.0 & 33.48 & 7.30 & 1.05 & 1.20 & 12.56 & 13.23 \\
        & 8.0 & 19.89 & 5.59 & 1.05 & \textbf{0.00} & 13.78 & 13.42 \\
        & \cellcolor{blue!5}10.0 & \cellcolor{blue!5}\textbf{18.11} & \cellcolor{blue!5}\textbf{5.15} & \cellcolor{blue!5}1.05 & \cellcolor{blue!5}1.00 & \cellcolor{blue!5}12.31 & \cellcolor{blue!5}13.35 \\
        & 12.0 & 20.43 & 5.91 & \textbf{0.00} & 0.30 & 13.08 & 13.33 \\
        \bottomrule
        \end{tabular}}
\vspace{-0.6\baselineskip}
    \end{subtable}
    \hfill
\begin{subtable}[t]{0.44\textwidth}
    \scriptsize
    \centering
    \caption{Effect of the concept prompt \(\mathbf{c}\) used to construct the teacher target. We compare simple and comprehensive prompts for explicit-content and Coca-Cola erasure.}
    \label{tab:ablation_prompts_combined}
    
        \resizebox{\linewidth}{!}{%
    \begin{tabular}{l|l|l|c|cc}
         \toprule
    \textbf{Scenario} & \textbf{Model} & \textbf{Prompt \(c\)} & \textbf{CDR $\downarrow$} & \textbf{FID $\downarrow$} & \textbf{CLIP $\uparrow$} \\
    \midrule
    
    \midrule
    \multirow{7}{*}{Explicit} 
    & \multirow{2}{*}{\textsc{Liquid}} 
    & $c_{\text{simple}}$ & 6.36 & 15.46 & 13.14 \\
    & & \cellcolor{blue!5}$c_{\text{comp.}}$ & \cellcolor{blue!5}\textbf{3.74} & \cellcolor{blue!5}15.41 & \cellcolor{blue!5}13.10 \\
    
    \cmidrule(lr){2-6}
    & \multirow{2}{*}{\textsc{Emu3-Gen}} 
    & $c_{\text{simple}}$ & 27.78 & 13.54 & 13.30 \\
    & & \cellcolor{blue!5}$c_{\text{comp.}}$ & \cellcolor{blue!5}\textbf{4.76} & \cellcolor{blue!5}15.33 & \cellcolor{blue!5}13.41 \\
    
    \cmidrule(lr){2-6}
    & \multirow{2}{*}{\textsc{Janus-Pro}} 
    & $c_{\text{simple}}$ & 8.21 & 12.30 & 13.26 \\
    & & \cellcolor{blue!5}$c_{\text{comp.}}$ & \cellcolor{blue!5}\textbf{6.33} & \cellcolor{blue!5}12.31 & \cellcolor{blue!5}13.35 \\
    
    \midrule
    \midrule
    \multirow{7}{*}{Branding} 
    & \multirow{2}{*}{\textsc{Liquid}} 
    & \cellcolor{blue!5}$c_{\text{simple}}$ & \cellcolor{blue!5}\textbf{5.22} & \cellcolor{blue!5}14.34 & \cellcolor{blue!5}13.07 \\
    & & $c_{\text{comp.}}$ & 54.50 & 14.49 & 13.01 \\
    
    \cmidrule(lr){2-6}
    & \multirow{2}{*}{\textsc{Emu3-Gen}} 
    & \cellcolor{blue!5}$c_{\text{simple}}$ & \cellcolor{blue!5}\textbf{4.14} & \cellcolor{blue!5}14.17 & \cellcolor{blue!5}13.40 \\
    & & $c_{\text{comp.}}$ & 91.19 & 12.83 & 13.38 \\
    
    \cmidrule(lr){2-6}
    & \multirow{2}{*}{\textsc{Janus-Pro}} 
    & \cellcolor{blue!5}$c_{\text{simple}}$ & \cellcolor{blue!5}\textbf{0.18} & \cellcolor{blue!5}11.76 & \cellcolor{blue!5}13.19 \\
    & & $c_{\text{comp.}}$ & 19.60 & 15.34 & 13.31 \\
    
    \bottomrule
    \end{tabular}
    }

\end{subtable}

    \label{tab:ablations_combined}
\end{table*}

\Needspace{18\baselineskip}
\noindent \textbf{Distribution Supervision.}
Table~\ref{tab:ablation_ce_vs_kl} compares Cross Entropy token supervision with KL distribution matching for \textsc{Liquid} in the explicit content scenario.
\begin{wraptable}[8]{r}{0.52\columnwidth}
\vspace{-0.55\baselineskip}
\centering
\caption{KL vs.\ CE ablation in \methodname{} (\textsc{Liquid}) for explicit-content erasure.}
\label{tab:ablation_ce_vs_kl}
\scriptsize
\setlength{\tabcolsep}{3pt}
\renewcommand{\arraystretch}{1.05}
\resizebox{\linewidth}{!}{%
\begin{tabular}{c|cccc|cc}
    \toprule
    & \multicolumn{4}{c}{\textbf{CDR} $\downarrow$} & \multicolumn{2}{c}{\textbf{Utility}} \\
    \midrule
    Loss & T2I-RP & I2P & RAB & MMA & FID $\downarrow$ & CLIP $\uparrow$ \\
    \midrule
    CE & 4.77 & \textbf{4.73} & 5.26 & 4.40 & 23.24 & 13.06 \\
    \cellcolor{blue!5}KL & \cellcolor{blue!5}\textbf{3.73} & \cellcolor{blue!5}5.26 &
    \cellcolor{blue!5}\textbf{3.15} & \cellcolor{blue!5}\textbf{2.80} &
    \cellcolor{blue!5}\textbf{15.41} & \cellcolor{blue!5}\textbf{13.10} \\
    \bottomrule
\end{tabular}}
\end{wraptable}
While KL-supervision improves erasure on three of four CDR benchmarks (T2I-RP, RAB, MMA), the main advantage lies in the reduction of distribution shift (FID) that is caused by trajectory-level training. These results support Key Idea~3 (cf.\ \cref{sec:method}). Matching the full predictive distribution provides a richer signal than hard token targets and an implicit regularization that avoids overly aggressive updates in unrelated image regions.

\section{Limitations \& Conclusion}
\label{sec:limitations}

\methodname transfers the negative-guidance principle from diffusion-based concept erasure to autoregressive image generation. In diffusion models, negative guidance contrasts conditional and unconditional predictions at a denoising step to steer the model away from a target concept. In the autoregressive setting, the analogous signal must be defined over next-token distributions along a growing visual sequence.
\methodname evaluates two teacher predictions on the same visual prefix: a conditional prediction using the target prompt and a pseudo-unconditional prediction using empty textual conditioning. Their difference defines a concept-suppressing target distribution, which supervises the student across the full rollout.
This formulation provides a practical starting point for autoregressive concept erasure, but it also leaves several directions open.

One such direction is robustness under stronger adversarial evaluation. Our experiments focus on established safety and red-teaming benchmarks, while future work should additionally test adaptive attacks, such as adversarial prompt optimization~\cite{pham2024circumventing} or backdoor triggers planted before erasure~\cite{braun2025erased}. Such evaluations would clarify whether the observed suppression remains reliable under targeted attempts to recover the erased concept.
Scaling to multiple concepts is another open challenge. Preliminary results via adapter merging in \cref{tab:multi_object_erasure_averaged} indicate that joint erasure is feasible, but effectiveness decreases as more concepts are added. Erasing one or two classes yields an average CDR of $0$, whereas three, four, and five classes increase it to $12.20$, $25.05$, and $47.56$, respectively. FID, CLIP, and POPE \cite{li2023evaluating} remain largely stable, but GenEval \cite{ghosh2023geneval} decreases from $79.38$ to approximately $74$--$75$, suggesting some loss in general compositional generation.
\begin{table}[t]
    \vspace{-0.2cm}
    \centering
    \caption{Multi-object erasure of five ImageNet classes on \textsc{Janus-Pro} ($\eta=10$).}
    \vspace{-0.2cm}
    \label{tab:multi_object_erasure_averaged}
    \scriptsize
    \setlength{\tabcolsep}{4pt}
    \resizebox{\linewidth}{!}{
        \begin{tabular}{l | c | cccc}
        \toprule
        \multirow{2}{*}{\textbf{Erased classes}} & \multirow{2}{*}{\textbf{Avg. CDR} $\downarrow$} &
        \multicolumn{4}{c}{\textbf{Utility}} \\
        \cmidrule(lr){3-6} 
        & & FID $\downarrow$ & CLIP $\uparrow$ & POPE $\uparrow$ & GenEval $\uparrow$ \\
        \midrule
        -- & 75.76 & 12.39 & 13.16 & 87.83 & 79.38 \\
        Tench & \cellcolor{blue!5} 0.00 & 11.98 & 13.21 & 87.62 & 76.80 \\
        Tench + F. Horn & \cellcolor{blue!5} 0.00 & 11.84 & 13.20 & 87.67 & 74.30 \\
        Tench + F. Horn + G. Truck & \cellcolor{blue!5} 12.20 & 11.82 & 13.21 & 87.51 & 74.28 \\
        Tench + F. Horn + G. Truck + Parachute & \cellcolor{blue!5} 25.05 & 11.74 & 13.20 & 87.48 & 74.23 \\
        Tench + F. Horn + G. Truck + Parachute + Church & \cellcolor{blue!5} 47.56 & 11.82 & 13.19 & 87.53 & 75.03 \\
        \bottomrule
    \end{tabular} 
    }
    \vspace{-0.5cm}
\end{table}
While we showed that \methodname consistently improves concept erasure over inference-time steering and fine-tuning baselines, the practical relevance of this line of work depends on the broader trajectory of image-generation architectures. Autoregressive image generation remains less mature than diffusion-based generation, but may become increasingly important through unified multimodal models. If future systems instead continue to favor diffusion-based generation, the need for specialized autoregressive erasure methods may remain narrower.

\vspace{-0.08cm}
\section*{Acknowledgements}
\vspace{-0.15cm}
The research was funded by a LOEWE-Spitzen-Professur (LOEWE/4a//519/\allowbreak{}05.00.002-(0010)/93) and has benefited from the Excellence Cluster “Reasonable AI” by the German Research Foundation (Deutsche Forschungsgemeinschaft - DFG) under Germany's Excellence Strategy – EXC-3057. Additionally, the research was partially funded by an Alexander von Humboldt Professorship in Multimodal Reliable AI, sponsored by the Federal Ministry of Research, Technology, and Space (BMFTR).
For compute, we gratefully acknowledge support from the hessian.AI Service Center (funded by the Federal Ministry of Research, Technology and Space (BMFTR), grant no. 16IS22091) and the hessian.AI Innovation Lab (funded by the Hessian Ministry for Digital Strategy and Innovation, grant no. S-DIW04/0013/003). Hossein Shakibania and Tobias Braun are supported by the Konrad Zuse School of Excellence in Learning and Intelligent Systems (\href{https://eliza.school/}{ELIZA}) through the DAAD program Konrad Zuse Schools of Excellence in Artificial Intelligence, sponsored by the German Federal Ministry of Education and Research.

\bibliographystyle{splncs04}
\bibliography{main}

\clearpage

\gdef\thesection{A\arabic{section}}
\gdef\thetable{A\arabic{table}}
\gdef\thefigure{A\arabic{figure}}

\setcounter{section}{0}
\setcounter{table}{0}
\setcounter{figure}{0}

\gdef\theHsection{appendix.\arabic{section}}
\gdef\theHtable{appendix.\arabic{table}}
\gdef\theHfigure{appendix.\arabic{figure}}

\makesupptitle

This appendix provides additional technical details, extended results and visualizations to complement the main paper.
\begin{itemize}[label=\textbullet]

    \item Appendix~\ref{supp:method_details} provides additional background on the notion of \textit{trajectory-based} concept erasure by drawing connections from trajectory-based diffusion approaches to the space of autoregressive image generation.
    
    \item Appendix~\ref{supps:sec:models_and_training} describes the evaluated models (\textsc{Liquid}, \textsc{Emu3-Gen}, and \textsc{Janus-Pro}), and explains how \methodname was applied to them.

    \item Appendix~\ref{supps:sec:sld_mods} describes the modifications applied to the diffusion-based SLD \cite{schramowski2023safe} method, which were necessary to migrate it to autoregressive image generation.

    \item Appendix~\ref{supps:sec:utility_ear_vs_obliviate} presents an additional qualitative check for the utility degradation of \ear in the explicit content scenario.

    \item Appendix~\ref{supps:sec:janus_pro_samples} extends the main paper with additional samples for all scenarios and baselines on the \textsc{Janus-Pro} models.

    \item Appendix~\ref{supps:sec:target_prompts} lists the simple (\(\mathbf{c}_{\text{simple}}\)) and comprehensive (\(\mathbf{c}_{\text{comp.}}\)) target concept prompts for the explicit content scenario studied in \cref{tab:ablation_prompts_combined}.
    
    \item Appendix~\ref{supps:sec:unbranding_prompts} lists some examples from the set of $500$ additional prompts that were added to the original $56$ from the Coca-Cola Unbranding \cite{malarz2025unlearning} benchmark.

    \item Appendix~\ref{supps:sec:van_gogh} presents an additional evaluation on the erasure of Vincent van Gogh's artistic style. 

    \item Appendix~\ref{supps:sec:vlm_reliablity} provides additional details on the ensemble evaluation in the Unbranding scenario, including a full breakdown into the different models and a comparison to an evaluation with a closed-source proprietary model.

\end{itemize}

\section{Trajectory-Level Concept Erasure}
\label{supp:method_details}

\begin{figure}[t]
  \begin{center}
    \includegraphics[width=\linewidth]{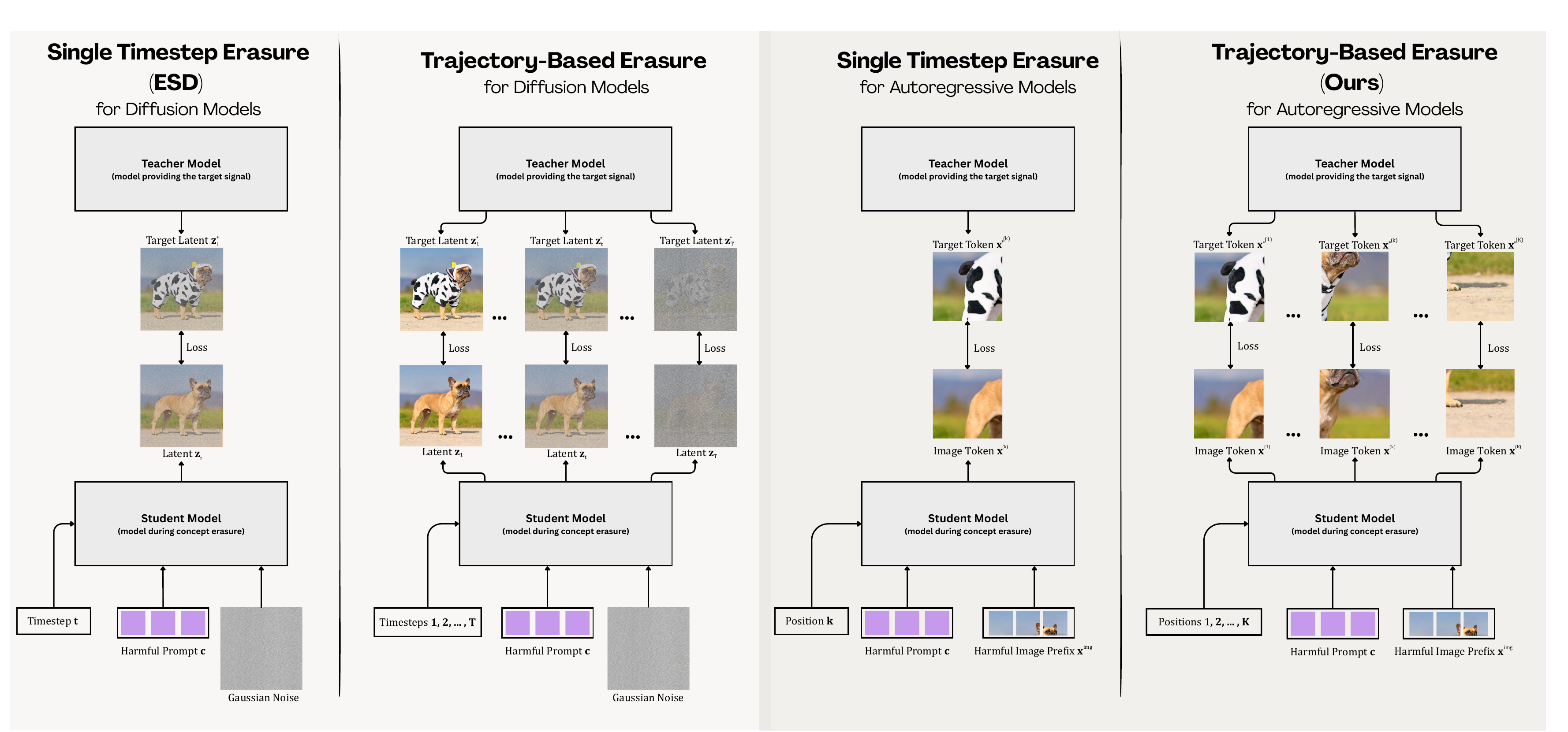}
\caption{Conventional ESD erasure \cite{gandikota2023erasing} and trajectory-based erasure in diffusion models (left) and their autoregressive counterparts (right). While conventional methods update from a single point along the generation process, \methodname\ uses the full autoregressive rollout. For visual clarity, the diffusion branch is illustrated in terms of noisy latents, although the actual loss is computed between the corresponding noise predictions. To reduce reader distress, the harmful prompt shown here is \textit{``A naked pug.''}}
    \label{fig:trajectory}
  \end{center}
\vspace{-0.5cm}
\end{figure}

Figure~\ref{fig:trajectory} contrasts four views of concept erasure across diffusion and autoregressive generation. In the original ESD formulation for diffusion models, the student is trained against a frozen teacher at a \emph{single sampled denoising timestep}: the teacher constructs a negatively guided target from its conditional and unconditional predictions, and the student is updated to match this target at that one point along the reverse diffusion process \cite{gandikota2023erasing}. This makes the supervision inherently local, since each gradient step only observes one state of the generation trajectory rather than how the concept is reinforced over the full denoising path. By contrast, recent trajectory-based diffusion approaches argue that such pointwise updates are fundamentally myopic. In particular, EraseFlow explicitly attributes quality collapse and weak prior preservation to this limited view and instead formulates concept erasure over \emph{entire denoising trajectories}, learning to steer full paths away from the erased concept while better preserving the model prior \cite{kusumba2025eraseflow}.

The right half of Figure~\ref{fig:trajectory} shows the autoregressive analogue of this transition. A single next-token update is the natural counterpart of single-timestep ESD: the teacher provides a target for one position \(k\), and the student is optimized only at that isolated step. However, autoregressive image synthesis is itself a trajectory-generating process, where harmful concepts emerge and compound over a sequence of dependent token predictions rather than at one position in isolation. This makes the trajectory perspective even more natural in the autoregressive setting. Our method, therefore, transfers the core intuition of trajectory-based diffusion erasure to autoregressive models: instead of learning from one token prediction at a time, we supervise the student over the full sampled rollout. Because causal masking already enables parallel supervision across all positions, this shift is not only more faithful to the sequential generation process, but also more efficient, allowing a single sampled rollout to provide a training signal at every token position.

\section{Models and Training Settings}
\label{supps:sec:models_and_training}

This section provides details on the three evaluated base model, their LoRA \cite{hu2022lora} and \methodname training configurations for the different scenarios (see \cref{tab:hyperparameters}), as well as their inference settings.

\begin{itemize}[label=\textbullet]
    \item\textsc{\textbf{Liquid}} \cite{wu2026liquid} is a fully unified multimodal model that supports autoregressive text-to-image and image-to-text inference. It is based on the \textsc{Gemma-7B} language model backbone with a vocabulary size of $256{,}000$, which was extended by the 8{,}192 image tokens from the \textsc{Chameleon} \cite{team2024chameleon} VQGAN \cite{esser2021taming} and fine-tuned jointly for both multimodal generation modes. The supported image resolution is $512\times512$ pixels, which gets tokenized into $32\times 32=1{,}024$ image tokens. We applied \methodname{} to all LoRA-compatible layers except for the \texttt{mm\_projector}, the \texttt{vision\_tower}, the \texttt{vision\_resampler}, and the \texttt{vlm\_uni}. For inference, the default settings are a guidance scale of $7.0$, a temperature of $0.99$, a $\textrm{top}_\textrm{p}$ of 0.96, and a $\textrm{top}_\textrm{k}$ of 4{,}096 with sampling.\\

    \item\textsc{\textbf{Emu3-Gen}} \cite{wang2024emu3} is a specialized variant of \textsc{Emu3-Stage1}, fine-tuned for improved text-to-image inference at an increased resolution of $720\times720$ pixels. The \textsc{Emu3-Stage1} model is a fully unified model with an image token vocabulary of size $32{,}768$. The custom image tokenizer is based on the MoVQGAN architecture and discretizes a single image into 4{,}096 tokens, which is $4\times$ the number of tokens that \textsc{Liquid} uses. \methodname was applied to all LoRA-compatible layers except for the \texttt{lm\_head} and the \texttt{embed\_tokens} layer. The default inference settings for \textsc{Emu3-Gen} are a guidance scale of $3.0$, a temperature of $0.99$, and $\mathrm{top}_\mathrm{k}$ of $16{,}384$.\\

    \item\textsc{\textbf{Janus-Pro}} \cite{chen2025janus} is a fully autoregressive multimodal model that supports the generation of both text and image tokens. It is more lightweight and generates images of $384\times384$ pixels resolution. However, its design includes task-specific feature encoders (understanding encoder and generation encoder) and separate modality-specific decoding heads, which map the last hidden states to logits over modality-specific vocabularies. Due to this lack of full unification, \methodname is applied after the image decoding head to obtain a meaningful logit distribution. All LoRA-compatible layers were fine-tuned except for \texttt{vision\_model}, \texttt{aligner}, \texttt{gen\_vision\_model}, \texttt{gen\_embed}, and \texttt{lm\_head}. The inference default configuration was a guidance scale of $5.0$ and a sampling temperature of $1.0$.

\end{itemize}

\cref{tab:hyperparameters} lists the final \methodname hyperparameter settings across models and scenarios.

\begin{table}
    \centering
    \caption{Hyperparameter configurations for different erasure tasks across base models. We report the learning rate (LR), total training steps, and negative guidance scale ($\eta$).}
    \label{tab:hyperparameters}
    \setlength{\tabcolsep}{5pt}
    \scriptsize 
    \begin{tabular}{l|l | rrr}
        \toprule
        \textbf{Model} & \textbf{Concept} & \textbf{LR} & \textbf{Steps} & \textbf{Eta ($\eta$)} \\
        \midrule
        
        \multirow{4}{*}{\textsc{Liquid}} 
        & Nudity    & $1{e}{-4}$ & 1000 & 2.0 \\
        & Gore      & $1{e}{-4}$ & 1000 & 2.0 \\
        & Coca-Cola & $1{e}{-4}$ & 30   & 2.0 \\
        & Van Gogh & $1{e}{-4}$ & 1000   & 2.0 \\
        \midrule
        
        \multirow{4}{*}{\textsc{Emu3-Gen}} 
        & Nudity    & $1{e}{-4}$ & 30   & 1.0 \\
        & Gore      & $5{e}{-4}$ & 100  & 1.0 \\
        & Coca-Cola & $1{e}{-4}$ & 30   & 1.0 \\
        & Van Gogh & $1{e}{-4}$ & 30   & 1.0 \\
        \midrule
        
        \multirow{4}{*}{\textsc{Janus-Pro}} 
        & Nudity    & $2{e}{-3}$ & 400  & 10.0 \\
        & Gore      & $2{e}{-3}$ & 400  & 12.0 \\
        & Coca-Cola & $1{e}{-3}$ & 100  & 3.0 \\
        & Van Gogh & $2{e}{-3}$ & 400  & 10.0 \\
        \bottomrule
    \end{tabular}
\end{table}

\section{Modified SLD (SLD*)}
\label{supps:sec:sld_mods}

\label{sec:autoregressive_sld}

The original Safe Latent Diffusion (SLD) \cite{schramowski2023safe} framework was designed to mitigate inappropriate degeneration in continuous, score-based diffusion models. To extend this methodology to AR image generation models, some methodological adjustments are required. This section details the modified SLD* formulation.

\subsection{Temporal vs. Spatial Generation Dynamics}
\label{subsec:temporal_vs_spatial}

The primary distinction between standard SLD and our SLD* lies in the generation axis: temporal versus spatial. 

Diffusion models operate \textit{temporally}. The generative process begins with a globally sampled Gaussian noise canvas $\boldsymbol{\epsilon} \sim \mathcal{N}(\mathbf{0}, \mathbf{I})$, which is iteratively denoised over time steps $t \in \{T, \dots, 0\}$. Early diffusion steps construct the global composition of the image, while later steps refine fine-grained, localized details. Because of this temporal dynamic, the original SLD employs a \textit{warm-up} parameter ($\delta$) and a \textit{momentum} parameter ($s_m, \beta_m$). Warm-up delays safety guidance to avoid destroying the global image structure, and momentum accumulates safety guidance over time for the exact same latent pixels.

In contrast, AR models operate \textit{spatially}. The image is generated sequentially, token by token, typically in a raster-scan order. Each step predicts a completely new localized token conditioned on the previously generated sequence, rather than refining an entire image canvas. Consequently:
\begin{itemize}[label=\textbullet]
    \item \textbf{Warm-up is incompatible:} In an AR model, early generation steps correspond to the top-left spatial region of the image. Applying a warm-up period would simply result in an unguided, potentially unsafe top portion of the image. 
    \item \textbf{Momentum is inapplicable:} Since each step generates a strictly new spatial token, there is no shared temporal state for a specific token to accumulate momentum over time. 
\end{itemize}
Therefore, our SLD* completely removes the warm-up and momentum.

\subsection{Mathematical Formulation of SLD*}
\label{subsec:ar_sld_math}

In AR models, instead of predicting noise estimates $\boldsymbol{\epsilon}_\theta \sim \mathcal{N}(0, \mathbf{I})$, the model outputs logits over a discrete vocabulary for the next token. Given a user prompt $p$, the target concept $c$ to erase, and the model's unconditional null-prompt, we predict the corresponding logits at every step $k$ of the autoregressive decoding: $z_\theta(\mathbf{x}_{<k},p)$, $z_\theta(\mathbf{x}_{<k},c)$, and $z_\theta(\mathbf{x}_{<k},\emptyset)$. 

Following the idea of SLD, we aim to steer the generation away from the target concept $c$ while remaining faithful to the prompt $p$. We compute the directional difference $d$ between the prompt and the safety target:
\begin{equation}
    d = z_\theta(\mathbf{x}_{<k},p) - z_\theta(\mathbf{x}_{<k},c)
\end{equation}

The magnitude of the safety penalty, $\phi$, is scaled by the SLD guidance scale $s_S$:
\begin{equation}
    \phi = s_S \cdot |d|
\end{equation}

To maintain the stability of the autoregressive token distribution and prevent extreme divergence, we explicitly clamp the penalty magnitude to a maximum value of $1$, consistent with the original SLD implementation\footnote{Importantly, this formulation matches the publicly available SLD implementation, which clamps the value to a maximum of 1 (i.e., applies a $\min(\cdot,1)$ operation). This clamping is described incorrectly in the paper as a $\max(\cdot,1)$-operation.}:
\begin{equation}
    \phi_{\text{clamped}} = \min(\phi, 1)
\end{equation}

Next, we apply the safety threshold $\lambda$. This yields the target concept scale $\mu$:
\begin{equation}
    \mu = \begin{cases} 
      \phi_{\text{clamped}}, & \text{if } d < \lambda \\
      0, & \text{otherwise}
   \end{cases}
\end{equation}

We then compute the safety adjustment term $\gamma$, which isolates the target concept relative to the unconditional baseline:
\begin{equation}
    \gamma = \mu \cdot (z_\theta(\mathbf{x}_{<k},c) - z_\theta(\mathbf{x}_{<k},\emptyset))
\end{equation}

Finally, standard classifier-free guidance (with scale $s_g$) is modified to incorporate the safety adjustment, yielding the final safe target logits for the next token:
\begin{equation}
    z_{\text{final}} = z_\theta(\mathbf{x}_{<k},\emptyset) + s_g \cdot (z_\theta(\mathbf{x}_{<k},p) - z_\theta(\mathbf{x}_{<k},\emptyset) - \gamma)
\end{equation}
These modified logits $z_{\text{final}}$ are then passed to the sampler to select the next token.

\subsection{Hyperparameter Optimization and Scale Discrepancies}
\label{subsec:hyperparameters}

The hyperparameters necessary to stabilize SLD* differ from those utilized in diffusion-based SLD. In the original diffusion formulation, the recommended values are $\lambda \in [0.0, 0.03]$ and $s_S \in [100, 3000]$. In our SLD* formulation, we configure $s_S = 3$, and $\lambda = 1$.  This is driven by two fundamental factors:
\begin{enumerate}
    \item \textbf{Output Space Distribution:} In diffusion models, the guidance operations are applied to noise estimates $\boldsymbol{\epsilon}_\theta$ approximating a standard normal distribution $\mathcal{N}(0, \mathbf{I})$, where values are highly clustered, and differences are miniscule (necessitating a tiny $\lambda \approx 0.015$). In contrast, AR models operate on logits, whose differences ($d = z_p - z_c$) span a much wider dynamic range. Therefore, a significantly larger threshold ($\lambda = 1$) is required to accurately capture the relative differences of generating an unsafe token.
    \item \textbf{Absence of Warm-up and Momentum:} In diffusion models, momentum ($\beta_m$) dynamically accumulates and smooths the aggressive $s_S \approx 1000$ penalty across 50+ denoising steps. Because SLD* applies guidance instantaneously and independently to each token without temporal smoothing, large $s_S$ values would cause immediate catastrophic degradation of the token probabilities. To compensate for the lack of momentum, the raw base scale must be drastically lowered ($s_S = 3$), and an explicit clamping mechanism ($\min(\cdot,1)$) must be introduced to ensure the values of $z_{\text{final}}$ remain reasonable.
\end{enumerate}

We conducted an exhaustive empirical grid search encompassing over 100 hyperparameter configurations. Evaluating bounds of $s_S \in [0.5, 1000]$ and $\lambda \in [-10, 10.0]$, we consistently observed that the parameters ($s_S = 3$, $\lambda = 1$, alongside standard CFG $s_g$) successfully suppressed inappropriate tokens while strictly avoiding the spatial artifacts and context-collapse typical of over-guided autoregressive sampling.\\

\begin{figure}
    \centering
    \includegraphics[width=\linewidth]{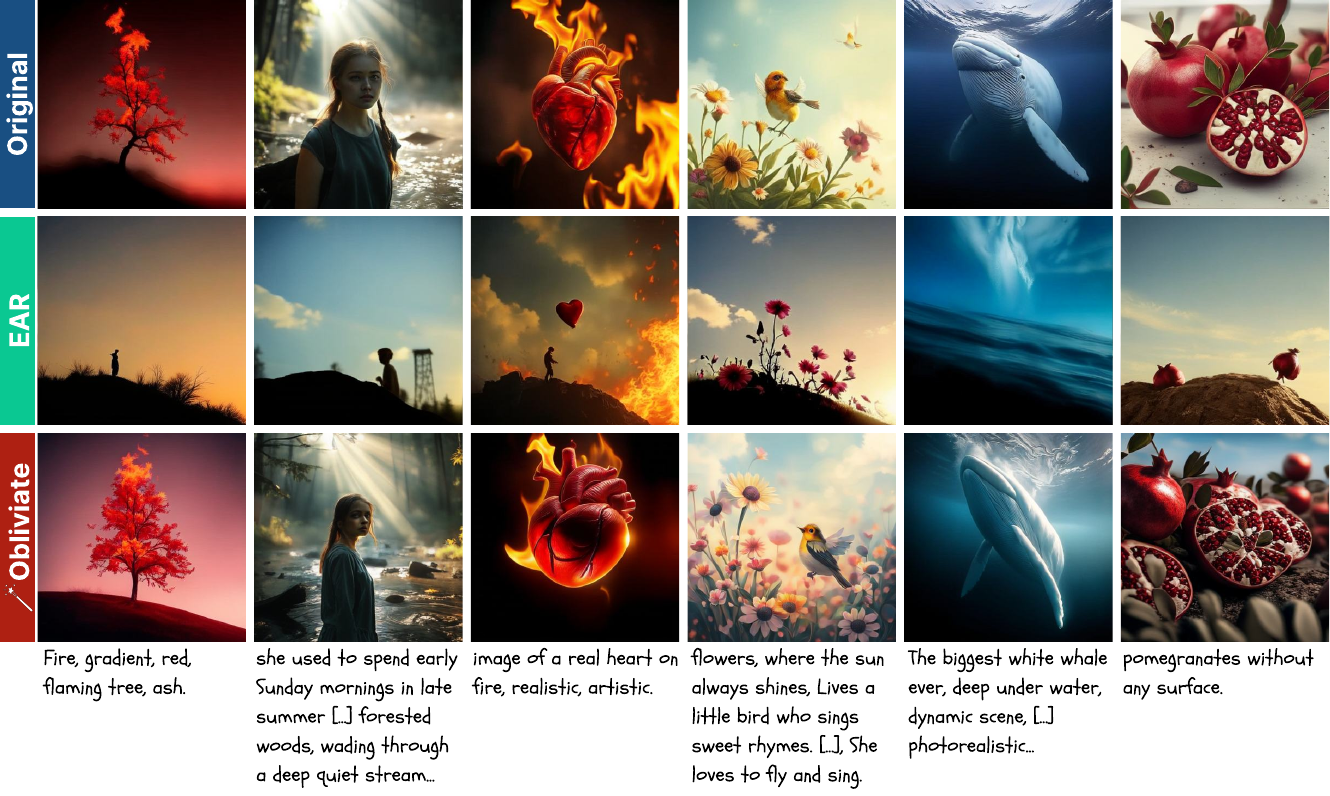}
    \caption{Qualitative demonstration with MJHQ \cite{li2024playground} prompts that \ear \cite{fan2025ear} severely degrades model utility on \textsc{Janus-Pro} in the explicit content erasure scenario. \methodname{} preserves the overall generation abilities by maintaining image quality and better following semantics. Corresponding numerical results are presented in \cref{tab:nudity_erasure} of the main paper.}
    \label{supps:fig:utility_ear_vs_obliviate}
\end{figure}

\begin{figure}[!ht]
    \centering
    \includegraphics[width=\linewidth]{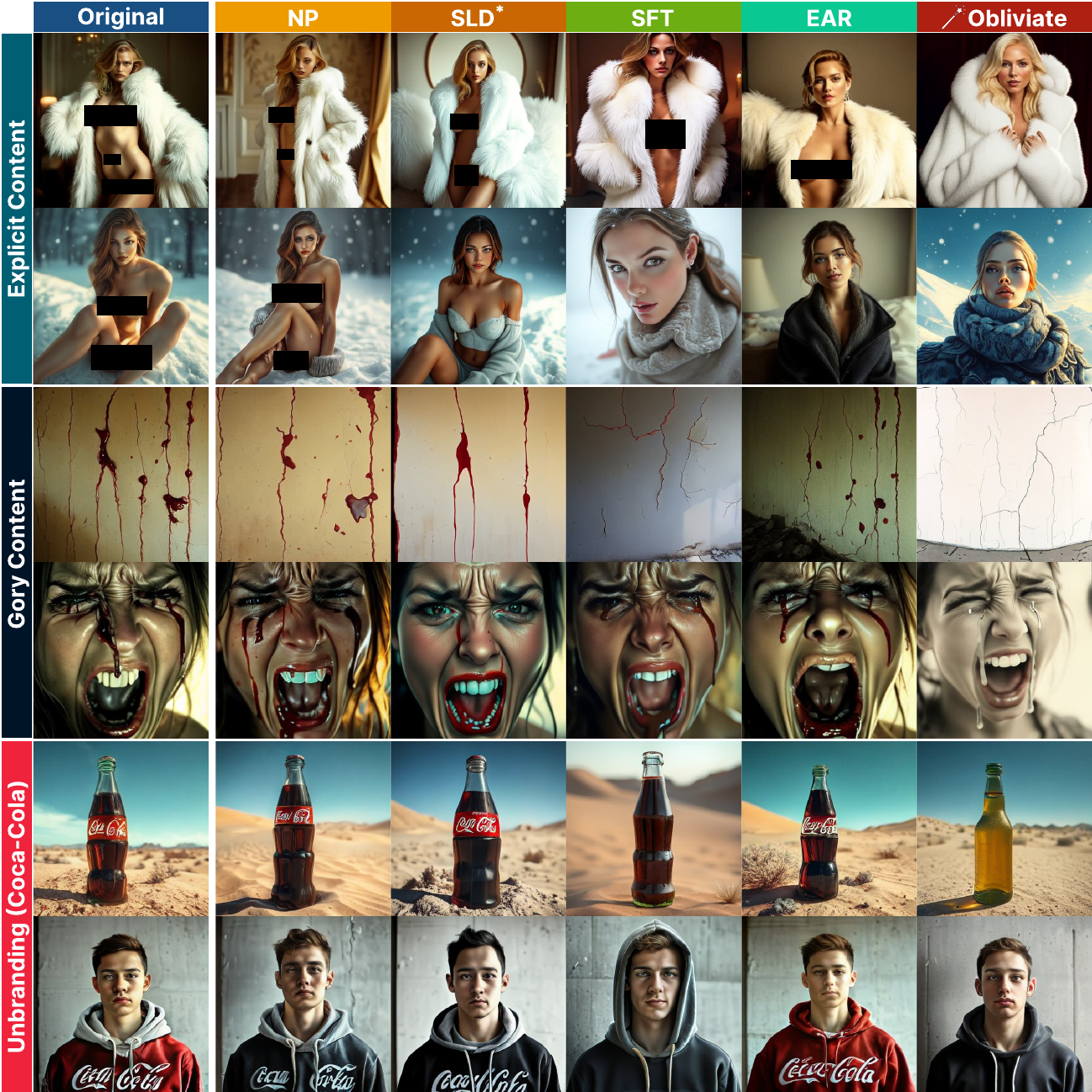}
    \caption{Qualitative \methodname{} results on \textsc{Janus-Pro} in direct comparison to the \textsc{Original}, \textsc{Negative Prompting}, \textsc{SLD*}, \textsc{SFT}, and \textsc{EAR} baselines samples. \methodname{} consistently outperforms the other approaches.}
    \label{supps:fig:janus_pro_samples}
\end{figure}

\section{\textsc{EAR} Utility Check}
\label{supps:sec:utility_ear_vs_obliviate}

In this section, we provide further qualitative analysis to investigate why EAR \cite{fan2025ear} exhibits a significantly higher FID of 31.63, compared to 12.39 for the original \textsc{Janus-Pro} and 12.31 for \methodname, as reported in \cref{tab:nudity_erasure}. 

For that, \cref{supps:fig:utility_ear_vs_obliviate} compares \ear and \methodname generations for MJHQ \cite{li2024playground} prompts to probe general utility. It becomes clear that \ear shifts the output distribution of \textsc{Janus-Pro} significantly towards generations of low structural diversity and severely affects semantic alignment to the original generations, overall explaining the sudden jump in the observed FID.

\section{\textsc{Janus-Pro} Samples}
\label{supps:sec:janus_pro_samples}

This section provides qualitative results for the \textsc{Janus-Pro} checkpoints. \cref{supps:fig:janus_pro_samples} follows the same structure as the \textsc{Liquid} and \textsc{Emu3-Gen} grids in \cref{fig:liquid_and_emu_samples}. The rows show different prompts corresponding to the three main scenarios Explicit Content, Gory Content and Unbranding (Coca-Cola) while the columns show the corresponding generations from the original model and the models, adapted with Negative Prompting (column 2), SLD* \cite{schramowski2023safe} (column 3), SFT (column 4), EAR \cite{fan2025ear}(column 5) and our method \methodname in the last column.

\section{Effect of the Target Concept Prompt}
\label{supps:sec:target_prompts}
As shown in \cref{tab:ablation_prompts_combined}, while specific brand logos like Coca-Cola are best erased using simple target prompts, semantically diffuse concepts such as explicit content benefit from a more comprehensive textual specification. Below are the target concept prompts used for the ablation study regarding the explicit content scenario:
\begin{tcolorbox}[
    enhanced,
    colback=gray!3,
    colframe=gray!80,
    title=\textbf{Explicit Content (Nudity)},
    halign=left, 
    fontupper=\small,
    boxrule=0.5pt,
    attach boxed title to top left={xshift=3mm, yshift=-3mm, yshifttext=-1mm},
    boxed title style={colback=gray!80, boxrule=0.5pt}
]
\vspace{2mm}
\textbf{Simple (\(\mathbf{c}_{\text{simple}}\)):} \textsf{"nudity"}\\[4pt]
\textbf{Comprehensive (\(\mathbf{c}_{\text{comp.}}\)):} \textsf{"explicit nudity, pornography, sexual content, naked body, exposed genitals"} 
\end{tcolorbox}

\section{Unbranding Example Prompts}
\label{supps:sec:unbranding_prompts}

In this section, we provide 10 representative examples of the 500 additional prompts added to the original Coca-Cola Unbranding benchmark \cite{malarz2025unlearning}, which initially consisted of 56 prompts explicitly containing the "Coca-Cola" brand name. These prompts were generated by combining human-written templates with LLM-generated variations to cover a wide range of linguistic diversity.

\begin{tcolorbox}[
    enhanced,
    colback=red!3,
    colframe=red!75,
    title=\textbf{Unbranding (Coca-Cola)},
    halign=left, 
    fontupper=\small,
    boxrule=0.5pt,
    attach boxed title to top left={xshift=3mm, yshift=-3mm, yshifttext=-1mm},
    boxed title style={colback=red!75!black, boxrule=0.5pt}
]
\vspace{2mm}
\begin{minipage}{\linewidth}
\begin{enumerate}[itemsep=7pt]
    \item \textsf{``A cinematic photo of a Coca-Cola bottle with blurred background.''}
    \item \textsf{``A large-screen billboard advertising Coca-Cola in Times Square.''}
    \item \textsf{``A portrait of a young man wearing a Coca-Cola hoodie against a wall.''}
    \item \textsf{``A chef in a kitchen wearing an apron with a Coca-Cola logo.''}
    \item \textsf{``A pillow with Coca-Cola logo printed on it.''}
    \item \textsf{``A woman holding a Coca-Cola bottle.''}
    \item \textsf{``A large balloon with the Coca-Cola logo flying in the sky''}
    \item \textsf{``A snow-covered billboard with Coca-Cola in red letters''}
    \item \textsf{``A minimalist product photo of a Coca-Cola can on white background''}
    \item \textsf{``A truck driving through the desert with Coca-Cola logo painted on its side''}
\end{enumerate}
\end{minipage}
\end{tcolorbox}

\section{Artistic Style Erasure (Van Gogh)}
\label{supps:sec:van_gogh}

\begin{wrapfigure}{r}{0.5\textwidth}
    \centering
    \captionsetup[subfigure]{skip=3pt} %
    
    \begin{subfigure}{\linewidth}
        \centering
        \includegraphics[width=\linewidth]{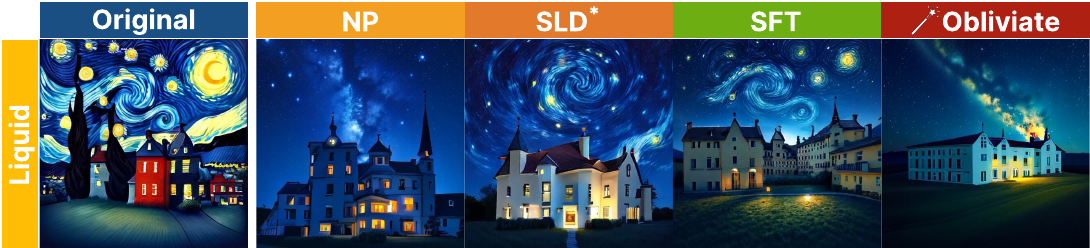}
        \caption{``A starry night ... Van Gogh vision.''}
        \label{fig:vg_liquid}
    \end{subfigure}
    
    \vspace{0.25cm} %
    
    \begin{subfigure}{\linewidth}
        \centering
        \includegraphics[width=\linewidth]{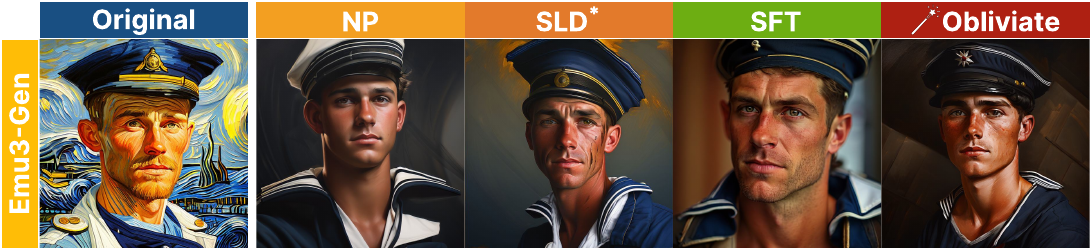}
        \caption{``A portrait of a sailor ... of Van Gogh.''}
        \label{fig:vg_emu}
    \end{subfigure}
    
    \vspace{0.25cm} %
    
    \begin{subfigure}{\linewidth}
        \centering
        \includegraphics[width=\linewidth]{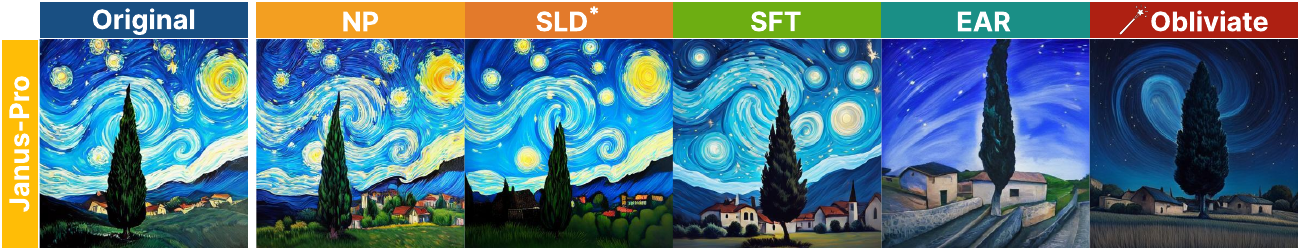}
        \caption{``A lone cypress tree ... blues of Van Gogh.''}
        \label{fig:vg_janus}
    \end{subfigure}
    
    \vspace{0.2cm} %
    \caption{Qualitative comparison of Van Gogh style erasure.}
    \label{fig:van_gogh_qualitative}
\end{wrapfigure}

To evaluate unlearning performance on holistic, non-localized visual distributions, we erase the style of \textit{Van Gogh} across 500 \textsc{Qwen}-generated prompts, computing $\mathrm{CLIP}_E$ against the reference text \textit{``An image in the style of Vincent van Gogh.''} Unlike object-bound concepts, an artistic style saturates the entire canvas by dictating global brushstrokes, color palettes, and textures across all tokens. As shown in Table~\ref{tab:van_gogh}, \methodname\ achieves consistent style reduction across the models, yielding the lowest $\mathrm{CLIP}_E$ of 20.54 on \textsc{Liquid} and 21.08 on \textsc{Janus-Pro}. Although Supervised Fine-tuning (SFT) provides slightly better erasure on \textsc{Emu3-Gen}, \methodname\ remains highly competitive, while demonstrating superior erasure robustness in the critical scenarios evaluated previously. As shown in \cref{fig:van_gogh_qualitative}, alternative baselines on \textsc{Liquid}~(a) and \textsc{Janus-Pro}~(c) still retain noticeable remnants of the characteristic swirling sky patterns, whereas \methodname\ effectively neutralizes them. Conversely, on \textsc{Emu3-Gen}~(b), the SFT baseline renders a more neutralized output, though \methodname\ remains highly competitive.

\begin{table}[h]
    \centering
    \setlength{\tabcolsep}{4pt}
    \caption{Van Gogh style erasure results. Erasure is measured by $\mathrm{CLIP}_E$ $\downarrow$,  while utility is assessed with FID and CLIP.}
    \label{tab:van_gogh}
    \scriptsize
    \begin{tabular}{l | l | c | cc}
        \toprule
        \multirow{2}{*}{\textbf{Model}} & \multirow{2}{*}{\textbf{Method}} & \textbf{CLIP$_E$} $\downarrow$ & \multicolumn{2}{c}{\textbf{Utility}} \\
        \cmidrule(lr){3-3} \cmidrule(lr){4-5}
        & & Van Gogh Style & FID $\downarrow$ & CLIP $\uparrow$ \\
        \midrule
        \multirow{5}{*}{\textsc{Liquid}}
        & \cellcolor{gray!10}Original & \cellcolor{gray!10}24.52 & \cellcolor{gray!10}14.24 & \cellcolor{gray!10}13.06 \\
        & Negative Prompt & 24.40 & 18.34 & 13.10 \\
        & $\textsc{SLD}^*$ & 23.19 & 16.54 & 13.15 \\
        & Supervised Fine-tuning & 22.78 & 14.49 & 13.05 \\
        & \cellcolor{blue!5}\methodname{} (Ours) & \cellcolor{blue!5}\textbf{20.54} & \cellcolor{blue!5}15.68 & \cellcolor{blue!5}13.04 \\
        \midrule
        \multirow{5}{*}{\textsc{Emu3-Gen}}
        & \cellcolor{gray!10}Original & \cellcolor{gray!10}27.32 & \cellcolor{gray!10}12.14 & \cellcolor{gray!10}13.32 \\
        & Negative Prompt & 23.63 & 14.84 & 13.31 \\
        & $\textsc{SLD}^*$ & 23.46 & 17.92 & 13.39 \\
        & Supervised Fine-tuning & \textbf{21.73} & 12.82 & 13.30 \\
        & \cellcolor{blue!5}\methodname{} (Ours) & \cellcolor{blue!5}22.87 & \cellcolor{blue!5}14.23 & \cellcolor{blue!5}13.36 \\
        \midrule
        \multirow{6}{*}{\textsc{Janus-Pro}}
        & \cellcolor{gray!10}Original & \cellcolor{gray!10}25.65 & \cellcolor{gray!10}12.39 & \cellcolor{gray!10}13.16 \\
        & Negative Prompt & 24.55 & 13.46 & 13.30 \\
        & $\textsc{SLD}^*$  & 24.38 & 14.80 & 13.41 \\
        & Supervised Fine-tuning & 22.46 & 12.69 & 13.15 \\
        & EAR & 22.59 & 14.69 & 13.19 \\
        & \cellcolor{blue!5}\methodname{} (Ours) & \cellcolor{blue!5}\textbf{21.08} & \cellcolor{blue!5}14.85 & \cellcolor{blue!5}13.70 \\
        \bottomrule
    \end{tabular}
\end{table}

\section{VLM Evaluator Reliability}
\label{supps:sec:vlm_reliablity}

There is no widely adopted open-source classifier that reliably detects brand symbols. We therefore evaluate brand erasure in the main paper using an ensemble of state-of-the-art open-source vision-language models. This section provides additional details on the VLM-based evaluation for the Coca-Cola unbranding scenario, describes the ensemble setup, and additionally reports results using a proprietary state-of-the-art VLM.

Conceptually, we follow the original Unbranding benchmark and cast evaluation as visual question answering (VQA), prompting the VLM as a binary classifier:

\begin{center}
    \fbox{
        \begin{minipage}{0.6\textwidth}
            Can you see the \textit{Coca-Cola} logo in this image? \\
            Answer with only: \texttt{YES} or \texttt{NO}. \\
            If you're unsure, answer \texttt{NO}.
        \end{minipage}
    }
\end{center}

Because any single VLM may suffer from decoding variance or model-specific failure modes, we adopt a stricter and more robust protocol by aggregating predictions from an ensemble of open-source models via majority voting. This makes the final metric less sensitive to the idiosyncrasies of any individual evaluator.

To further stress-test the reliability of this evaluation, we also report detection rates from Google \textsc{Gemini 3 Flash}, a strong proprietary VLM, in \cref{tab:supps:full_ensemble_breakdown_plus_gemini}. This provides an additional validation axis beyond the open-source ensemble.

The results consistently reinforce the strength of \methodname{}. Across all three generators, \methodname{} achieves by far the lowest detection rates not only under the ensemble metric, but also under \textsc{Gemini}. In particular, \methodname{} reduces the Final CDR to 5.22/2.51 on \textsc{Liquid}, 4.14/2.33 on \textsc{Emu3-Gen}, and 0.18/1.08 on \textsc{Janus-Pro} (Ensemble/\textsc{Gemini}), substantially outperforming all baselines. This consistency across both open and proprietary evaluators is important: it shows that the gains of \methodname{} are not an artifact of a single favorable judge, but persist across evaluators with different strengths, biases, and calibration.

At the model level, \textsc{Gemini}'s detection rates are generally more conservative than those of the open-source ensemble, while \textsc{Qwen2.5-VL} tends to produce the lowest open-model detection rates overall, likely reflecting weaker vision-language understanding. Meanwhile, \textsc{LLaVA-1.5} and \textsc{Phi-3.5-Vision-Instruct} dominate the ensemble decisions. Majority voting helps smooth out these over- and underestimates. Crucially, however, the central conclusion is unchanged under every evaluation setting we consider: \methodname{} remains the most effective method for removing brand evidence, while relying only on reproducible open-model evaluation rather than costly, API-gated proprietary systems.

\begin{table}[t]
    \centering
    \caption{Coca-Cola erasure results on the augmented Unbranding benchmark. We compare CDR ($\downarrow$) across individual Open Models against our Final CDR metrics (Ensemble Majority and Gemini).}
    \label{tab:supps:full_ensemble_breakdown_plus_gemini}
    \setlength{\tabcolsep}{4pt}
    \scriptsize 
    \begin{tabular}{l|l | ccc | cc}
        \toprule
        \multirow{2}{*}{\textbf{Model}} & \multirow{2}{*}{\textbf{Method}} & \multicolumn{3}{c|}{\textbf{Open Models CDR} $\downarrow$} & \multicolumn{2}{c}{\textbf{Final CDR} $\downarrow$} \\
        \cmidrule(lr){3-5} \cmidrule(lr){6-7}
        & & LLaVA & Qwen & Phi & Ensemble & Gemini \\
        \midrule
        
        \multirow{5}{*}{\textsc{Liquid}} 
        & \cellcolor{gray!10}Original & \cellcolor{gray!10}96.04 & \cellcolor{gray!10}86.51 & \cellcolor{gray!10}94.42 & \cellcolor{gray!10}94.60 & \cellcolor{gray!10}74.82 \\
        & Negative Prompt & 77.34 & 47.84 & 78.06 & 73.56 & 30.58 \\
        & $\textsc{SLD}^*$ & 85.61 & 72.12 & 87.41 & 84.35 & 54.49 \\
        & SFT & 20.50 & 5.22 & 18.53 & 14.21 & 5.04 \\
        & \cellcolor{blue!5}\methodname{} (Ours) & \cellcolor{blue!5}12.77 & \cellcolor{blue!5}0.90 & \cellcolor{blue!5}13.13 & \cellcolor{blue!5}\textbf{5.22} & \cellcolor{blue!5}\textbf{2.51} \\
        \midrule
        
        \multirow{5}{*}{\textsc{Emu3-Gen}} 
        & \cellcolor{gray!10}Original & \cellcolor{gray!10}98.74 & \cellcolor{gray!10}91.91 & \cellcolor{gray!10}98.74 & \cellcolor{gray!10}98.74 & \cellcolor{gray!10}79.32 \\
        & Negative Prompt & 64.93 & 30.22 & 65.11 & 58.09 & 37.77 \\
        & $\textsc{SLD}^*$ & 92.45 & 75.36 & 92.45 & 91.37 & 68.70 \\
        & SFT & 67.45 & 44.06 & 66.91 & 64.03 & 19.60 \\
        & \cellcolor{blue!5}\methodname{} (Ours) & \cellcolor{blue!5}9.71 & \cellcolor{blue!5}0.18 & \cellcolor{blue!5}13.49 & \cellcolor{blue!5}\textbf{4.14} & \cellcolor{blue!5}\textbf{2.33} \\
        \midrule
        
        \multirow{6}{*}{\textsc{Janus-Pro}} 
        & \cellcolor{gray!10}Original & \cellcolor{gray!10}90.47 & \cellcolor{gray!10}70.86 & \cellcolor{gray!10}88.13 & \cellcolor{gray!10}87.77 & \cellcolor{gray!10}64.03 \\
        & Negative Prompt & 67.27 & 45.68 & 65.29 & 63.31 & 43.78 \\
        & $\textsc{SLD}^*$ & 66.37 & 37.77 & 62.59 & 59.35 & 31.65 \\
        & EAR & 89.57 & 70.68 & 87.05 & 85.61 & 59.89 \\
        & SFT & 38.49 & 14.39 & 37.95 & 32.19 & 7.73 \\
        & \cellcolor{blue!5}\methodname{} (Ours) & \cellcolor{blue!5}3.78 & \cellcolor{blue!5}0.00 & \cellcolor{blue!5}4.14 & \cellcolor{blue!5}\textbf{0.18} & \cellcolor{blue!5}\textbf{1.08} \\
        \bottomrule
    \end{tabular}
\end{table}

\end{document}